\pdfoutput=1
\documentclass[11pt]{article}

\usepackage[preprint]{acl}

\usepackage{times}
\usepackage{latexsym}
\usepackage[T1]{fontenc}
\usepackage[utf8]{inputenc}

\usepackage{microtype}
\usepackage{graphicx}
\usepackage{tikz}
\usetikzlibrary{arrows.meta,positioning,fit,backgrounds,calc,shapes.geometric,shapes.symbols,decorations.pathreplacing}
\graphicspath{{./}}
\usepackage{booktabs}
\usepackage{multirow}
\usepackage{amsmath}
\usepackage{amssymb}
\usepackage{amsthm}
\usepackage{bm}
\usepackage{xcolor}
\usepackage{array}
\makeatletter
\@ifundefined{insert@pcolumn}{\let\insert@pcolumn\insert@column}{}
\makeatother
\usepackage{colortbl}
\usepackage{pifont}
\usepackage{makecell}
\usepackage{placeins}
\usepackage{tcolorbox}
\tcbuselibrary{breakable,skins}
\usepackage{xurl}
\usepackage{subcaption}
\usepackage{fontawesome5}
\usepackage{xspace}
\usepackage[capitalize,nameinlink,noabbrev]{cleveref}

\definecolor{ink}{HTML}{111827}
\definecolor{muted}{HTML}{5B6474}
\definecolor{hpurple}{HTML}{5445B5}
\definecolor{hindigo}{HTML}{211CBF}
\definecolor{lavtint}{HTML}{ECECFC}
\definecolor{boxback}{HTML}{F5F5FE}
\definecolor{boxframe}{HTML}{A7A8F0}
\definecolor{boxtitle}{HTML}{8180DE}
\definecolor{paletteCyan}{HTML}{8FD2EC}
\definecolor{paletteMagenta}{HTML}{D38FCC}
\definecolor{paletteP1}{HTML}{ADAFF4}
\definecolor{paletteP2}{HTML}{989EEC}
\definecolor{paletteP3}{HTML}{8482E6}
\definecolor{paletteP4}{HTML}{6962D2}
\definecolor{paletteP5}{HTML}{5246BC}
\definecolor{rawBlueLight}{HTML}{DDF2FD}
\definecolor{qualMagentaLight}{HTML}{FBE3FF}
\definecolor{cleanPurpleLight}{HTML}{E6E6FA}
\definecolor{robustPurpleMed}{HTML}{CFCFF5}

\newcommand{\cmark}{\ding{51}}
\newcommand{\xmark}{\ding{55}}

\newcommand{\eg}{\textit{e.g.}\xspace}

\newcommand{\vs}{\textit{vs.}\@\xspace}

\DeclareRobustCommand{\glogo}{\raisebox{-0.27\height}{\includegraphics[height=2.35ex]{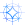}}\hspace{0.22em}}
\DeclareRobustCommand{\gtwob}{\glogo Gemma~2 2B}
\DeclareRobustCommand{\gninb}{\glogo Gemma~2 9B}
\DeclareRobustCommand{\gthrb}{\glogo Gemma~3 4B}
\DeclareRobustCommand{\gtwoS}{\glogo 2B}
\DeclareRobustCommand{\gninS}{\glogo 9B}
\DeclareRobustCommand{\gthrS}{\glogo 3-4B}

\crefname{section}{\S}{\S\S}
\Crefname{section}{\S}{\S\S}
\crefname{appendix}{\S}{\S\S}
\Crefname{appendix}{\S}{\S\S}
\crefformat{section}{\S#2#1#3}
\crefformat{subsection}{\S#2#1#3}
\crefformat{appendix}{\S#2#1#3}
\crefformat{subappendix}{\S#2#1#3}
\crefmultiformat{section}{\S#2#1#3}{ and~\S#2#1#3}{, \S#2#1#3}{ and~\S#2#1#3}
\crefmultiformat{subsection}{\S#2#1#3}{ and~\S#2#1#3}{, \S#2#1#3}{ and~\S#2#1#3}
\crefmultiformat{appendix}{\S#2#1#3}{ and~\S#2#1#3}{, \S#2#1#3}{ and~\S#2#1#3}
\crefmultiformat{subappendix}{\S#2#1#3}{ and~\S#2#1#3}{, \S#2#1#3}{ and~\S#2#1#3}
\Crefformat{section}{\S#2#1#3}
\Crefformat{subsection}{\S#2#1#3}
\Crefformat{appendix}{\S#2#1#3}
\Crefformat{subappendix}{\S#2#1#3}
\crefname{figure}{Figure}{Figures}
\crefname{table}{Table}{Tables}
\crefname{equation}{Eq.}{Eqs.}
\crefname{proposition}{Proposition}{Propositions}
\crefname{remark}{Remark}{Remarks}
\crefname{tcb@cnt@numberedbox}{Box}{Boxes}

\theoremstyle{plain}
\newtheorem{proposition}{Proposition}
\theoremstyle{remark}
\newtheorem{remark}{Remark}

\newcommand{\squishlist}{
\begin{list}{$\bullet$}
{   \setlength{\itemsep}{0pt}
   \setlength{\parsep}{2pt}
   \setlength{\topsep}{2pt}
   \setlength{\partopsep}{0pt}
   \setlength{\leftmargin}{1.2em}
   \setlength{\labelwidth}{0.8em}
   \setlength{\labelsep}{0.4em} } }
\newcommand{\squishend}{\end{list} }

\newtcolorbox[auto counter, number within=section]{numberedbox}[2][]{%
  breakable, enhanced, colback=boxback, colframe=boxframe, colbacktitle=boxtitle, coltitle=white,
  boxrule=0.5pt, arc=2pt, left=5pt, right=5pt, top=3pt, bottom=3pt, fonttitle=\bfseries\small,
  fontupper=\small, title={Box \thetcbcounter: #2}, #1}

\definecolor{tocgist}{HTML}{5B6474}
\newcommand{\appentry}[3]{%
  \noindent\hyperref[#1]{\makebox[1.25em][l]{\ref*{#1}}#2}\dotfill\pageref{#1}\par
  \vspace{-0.5pt}{\leftskip=1.25em\noindent\color{tocgist}\fontsize{8.6}{10}\selectfont #3\par}\vspace{3pt}}

\title{SAE-StatSteer: Statistical Consensus Feature Selection for Optimization-Free Activation Steering of Large Language Models}

\author{\bf Oshayer Siddique$^*$,
{\bf J. M Areeb Uzair Alam$^*$,}
{\bf Md Jobayer Rahman Rafy$^*$,}\\
{\bf Syed Rifat Raiyan$^{*\text{\textdagger}}$,}
{\bf Hasan Mahmud,}
{\bf Md Kamrul Hasan}\\
Systems and Software Lab (SSL)\\Department of Computer Science and Engineering\\
Islamic University of Technology, Dhaka, Bangladesh\\
\texttt{\small\{oshayer, areebuzair, jobayerrahman, rifatraiyan, hasan, hasank\}@iut-dhaka.edu}\\
\small\faGithub\;\href{https://github.com/Oshayer-Siddique/SAE-STATSTEER}{\texttt{Oshayer-Siddique/SAE-STATSTEER}}
}

\begin{document}
\maketitle
{\def\thefootnote{$*$}\footnotetext{Equal contribution.}}
{\def\thefootnote{\textdagger}\footnotetext{Corresponding author.}}

\begin{abstract}
Activation steering adds a residual-stream direction at inference time, providing lightweight behavioral control without fine-tuning. Sparse autoencoders (SAEs) can make such interventions auditable by decomposing dense activations into an approximately monosemantic feature basis. We introduce SAE-StatSteer, a transparent, optimization-free pipeline. It filters features through six reliability conditions, ranks the survivors by an unweighted Borda consensus over three statistics (an $F$-test, KSG mutual information, and Cohen's $d$), and combines the selected SAE decoder rows using Cohen's-$d$ weights. We evaluate three Gemma-family models across four behavioral domains against seven dense or SAE-based baselines, under a quality-conditioned protocol that requires attribute movement while preserving relevance, richness, and coherence. Within this Gemma scope, raw success systematically overstates usable control because strong shifts often degrade generation quality, and effective steering is not governed by a universal layer or strength. SAE-StatSteer reaches parity with optimization-based methods while exposing every selection and weighting decision for audit, and removing its consensus ranking or its reliability filter at least halves clean success. These results motivate reporting quality-conditioned success alongside raw behavioral shift.
\end{abstract}

\begin{figure}[!t]
  \centering
  \resizebox{\columnwidth}{!}{%
  \begin{tikzpicture}[
    font=\scriptsize,
    node distance=4mm and 5mm,
    stage1/.style={rectangle, draw=paletteP4, rounded corners=2pt, minimum height=7mm, minimum width=22mm, align=center, fill=paletteCyan!35, line width=0.5pt},
    stage2/.style={rectangle, draw=paletteP4, rounded corners=2pt, minimum height=6mm, minimum width=20mm, align=center, fill=paletteMagenta!30, line width=0.5pt},
    stage3/.style={rectangle, draw=paletteP5, rounded corners=2pt, minimum height=6mm, minimum width=24mm, align=center, fill=paletteP1!55, line width=0.5pt},
    test/.style={rectangle, draw=paletteP4, rounded corners=2pt, minimum height=5mm, minimum width=14mm, align=center, fill=paletteCyan!50, line width=0.4pt, font=\tiny},
    arr/.style={-{Stealth[length=4pt]}, line width=0.45pt, color=paletteP4!85!black},
  ]
    \node[stage1] (pairs) {800 contrast pairs\\per domain $D$};
    \node[stage1, right=of pairs] (extract) {SAE encode\\max-pool $\to Z_D$};
    \node[test, below=8mm of pairs.south west, anchor=north west, xshift=-2mm] (fstat) {$F$-test};
    \node[test, right=2mm of fstat] (mi) {MI (KSG)};
    \node[test, right=2mm of mi] (cohen) {Cohen's $d$};
    \node[stage2, below=4mm of mi] (filter) {6-condition quality filter};
    \node[stage2, below=4mm of filter] (borda) {Borda consensus\\(3/3 then 2/3 tier)};
    \node[stage2, below=4mm of borda] (topk) {Select top-$K\!\in\!\{16,24,32\}$:\\indices $\mathbf{f}$, weights $\mathbf{d}$};
    \node[stage3, right=8mm of topk.east, anchor=west] (vinit) {$\mathbf{v}_{\text{init}}{=}\mathbf{d}/\|\mathbf{d}\|$};
    \node[stage3, above=4mm of vinit] (dh) {$\delta\mathbf{h}{=}\mathbf{v}_{\text{init}}^\top W_{\text{dec}}[\mathbf{f},:]$};
    \node[stage3, above=4mm of dh] (inject) {Hook at layer $\ell$:\\$h{\leftarrow}h{+}\alpha\,\|h\|\,\hat{\delta\mathbf{h}}$};
    \draw[arr] (pairs) -- (extract);
    \draw[arr] (extract.south) |- ($(fstat.north)+(0,4mm)$) -| (fstat.north);
    \draw[arr] (extract.south) |- ($(mi.north)+(0,4mm)$) -| (mi.north);
    \draw[arr] (extract.south) |- ($(cohen.north)+(0,4mm)$) -| (cohen.north);
    \draw[arr] ($(fstat.south)+(-3mm,0)$) |- ($(filter.west)+(-3mm,0)$) -- (filter.west);
    \draw[arr] (mi) -- (filter);
    \draw[arr] ($(cohen.south)+(3mm,0)$) |- ($(filter.east)+(3mm,0)$) -- (filter.east);
    \draw[arr] (filter) -- (borda);
    \draw[arr] (borda) -- (topk);
    \draw[arr] (topk.east) -- (vinit.west);
    \draw[arr] (vinit) -- (dh);
    \draw[arr] (dh) -- (inject);
    \begin{pgfonlayer}{background}
      \node[draw=paletteCyan!80!black, dashed, rounded corners, fit=(pairs)(extract), inner sep=2.5pt, label={[font=\tiny, color=paletteP4]above:\textbf{Stage 1: Extract}}] {};
      \node[draw=paletteMagenta!85!black, dashed, rounded corners, fit=(fstat)(cohen)(filter)(borda)(topk), inner xsep=2.5pt, inner ysep=8pt, label={[font=\tiny, color=paletteP4]above:\textbf{Stage 2: Select}}] {};
      \node[draw=paletteP5, dashed, rounded corners, fit=(vinit)(dh)(inject), inner sep=2.5pt, label={[font=\tiny, color=paletteP5]above:\textbf{Stage 3: Steer}}] {};
    \end{pgfonlayer}
  \end{tikzpicture}}
  \caption{Compact teaser of the three-stage pipeline: contrastive SAE feature extraction, consensus-ranked feature selection, and Cohen's-$d$-weighted residual-stream steering.}
  \label{fig:pipeline}
  \vspace{-1mm}
\end{figure}

\section{Introduction}
\label{sec:intro}

A steering vector is easy to inject and hard to read. Activation steering adds a direction to an internal residual stream at inference time and can shift a model's behavior without updating a single weight \citep{turner2024actadd,rimsky2024caa,zou2023repr}, yet a dense direction may entangle the intended behavior with topic, style, or formatting artifacts. Sparse autoencoders (SAEs) offer a more inspectable basis by decomposing dense activations into sparse features that are often approximately monosemantic \citep{bricken2023monosemanticity,templeton2024scaling,lieberum2024gemmascope}. A steering vector assembled from named SAE features can therefore be inspected feature by feature rather than treated as an unexplained dense perturbation.

Existing SAE-steering methods establish that this approach is viable, but they introduce different dependencies. SAE-SSV trains probes and optimizes a steering vector in the selected subspace \citep{he2025saessv}; SAE-RSV uses an LLM judge to refine feature semantics \citep{wang2025saersv}; and SAE-Steering combines keyword-based recall with intervention ranking \citep{fang2026saesteering}. These methods motivate a simpler question: how far can a transparent, optimization-free selector go when every aggregation decision is stated explicitly? A statistical answer is attractive because feature evidence can be compared, audited, and reproduced without the tacit choices that learned aggregation weights or a gradient-based vector objective entail.

\begin{figure*}[t]
    \centering
    \includegraphics[width=\linewidth]{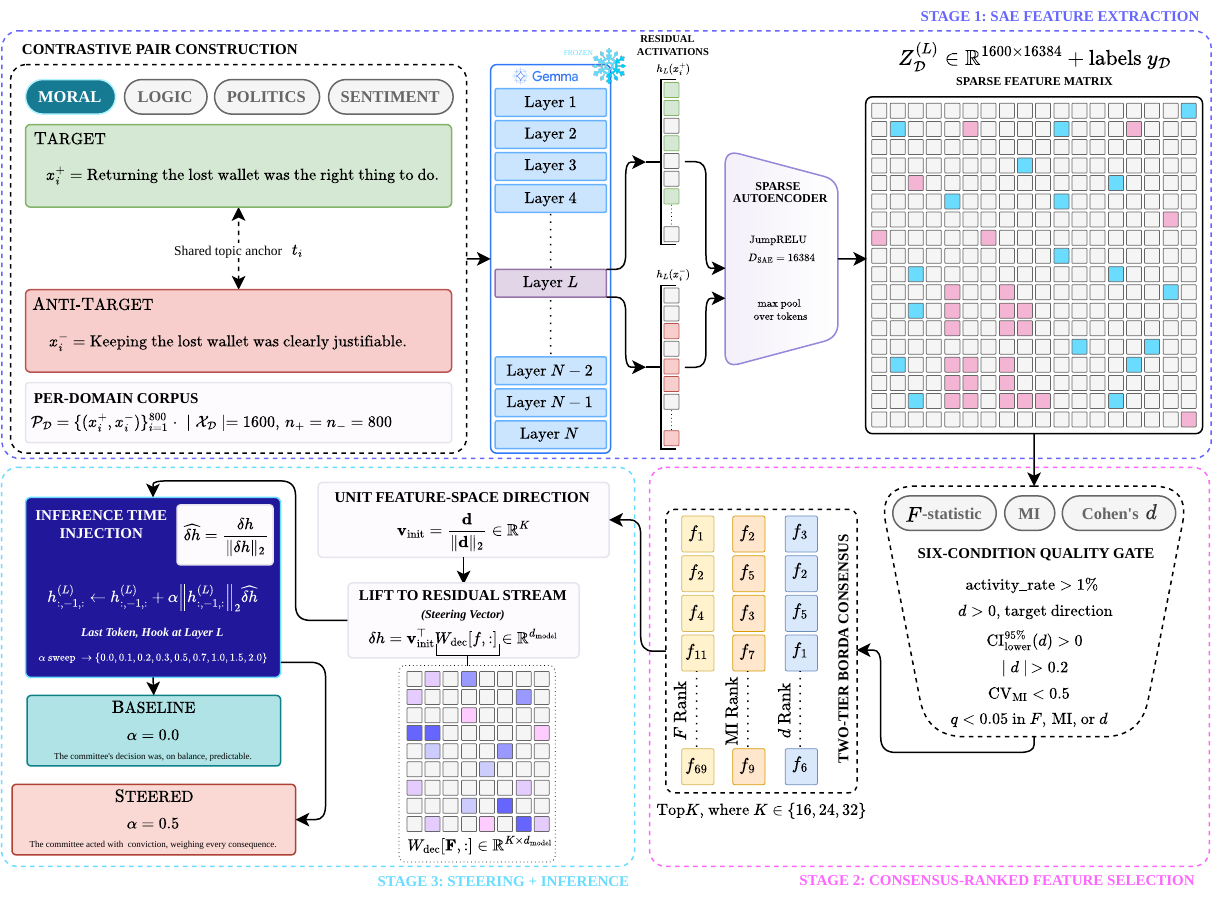}
    \caption{A detailed view of the pipeline sketched in \cref{fig:pipeline}. \textbf{Stage~1 (SAE feature extraction):} for each domain, $800$ topic-anchored contrastive pairs pass through the frozen model; post-MLP residual activations at a chosen layer $\ell$ are encoded by a JumpReLU SAE ($D_{\mathrm{SAE}}{=}16{,}384$) and max-pooled over tokens into the sparse feature matrix $Z_{D}^{(\ell)}\in\mathbb{R}^{1600\times16384}$ with labels $y_{D}$. \textbf{Stage~2 (consensus-ranked feature selection):} every feature is scored by three complementary statistics ($F$-test, KSG mutual information, and Cohen's $d$), filtered by the six-condition reliability gate, and ranked by an unweighted two-tier Borda consensus; candidate sets of $K\in\{16,24,32\}$ features (or the whole consensus pool, if smaller) yield indices $\mathbf{f}$ and weights $\mathbf{d}$, and each steered layer uses one of them. \textbf{Stage~3 (steering and inference):} the unit feature-space direction $\mathbf{v}_{\mathrm{init}}{=}\mathbf{d}/\lVert\mathbf{d}\rVert$ is lifted to the residual stream as $\delta\mathbf{h}{=}\mathbf{v}_{\mathrm{init}}^{\top}W_{\mathrm{dec}}[\mathbf{f},:]$ and injected at the last-token position with activation-proportional scaling, $h^{(\ell)}_{:,-1,:}\leftarrow h^{(\ell)}_{:,-1,:}+\alpha\,\lVert h^{(\ell)}_{:,-1,:}\rVert_{2}\,\widehat{\delta\mathbf{h}}$, swept over $\alpha\in\{0,0.1,0.2,0.3,0.5,0.7,1.0,1.5,2.0\}$ against an $\alpha{=}0$ baseline.}
    \label{fig:main_pipeline}
\end{figure*}

Two evaluation gaps are equally germane. Prior work has observed that steering trades target movement against generation quality and that its operating point needs case-by-case tuning \citep{wu2025axbench,stolfo2025improving}; what is missing is a systematic measurement of both effects. Studies commonly report a single intervention layer or a narrow strength range, although steering behavior can vary sharply across depth, model scale, and target attribute. Raw attribute movement is also not the same as useful control: a large shift can come from text that is repetitive, incoherent, or off topic, which a quality-blind evaluation would still reward. A practical criterion therefore has two desiderata: it should require target movement together with preserved relevance, richness, and coherence, and it should survey enough layers and strengths to test whether any operating point generalizes.

We introduce SAE-StatSteer (\cref{fig:pipeline}). For each behavioral domain and candidate layer, it extracts SAE activations from topic-matched contrast pairs, filters unreliable features, and combines three rankings with an unweighted Borda consensus: an $F$-test, KSG mutual information \citep{kraskov2004ksg}, and Cohen's $d$ \citep{cohen1988statistical}. Selected decoder rows are weighted by their signed Cohen's-$d$ values, a weighting midway between the mean difference and diagonal Fisher LDA. We test three Gemma-family models on four domains and compare four dense and three SAE-based steering baselines under one three-judge, quality-conditioned protocol. Our contributions are as follows:

\squishlist
    \item We propose a transparent SAE steering pipeline that combines a six-condition reliability filter, three-statistic Borda consensus, and a closed-form, Cohen's-$d$-weighted decoder-row direction, with neither learned aggregation weights nor gradient optimization of the vector; an ablation shows that the consensus and the filter each carry a large share of its clean success.

    \item We evaluate three model scales and four behavioral domains across 344 single-layer settings and 12 multi-layer mixtures, counting a success only when the target moves while relevance, richness, and coherence survive.

    \item We quantify two effects that prior work observed without measuring them across a sweep: raw success exceeds clean, usable success by 4.1 to 28.0 percentage points per cell, and neither relative layer depth nor steering strength relates monotonically to target movement.

    \item We compare seven dense or SAE-based baselines under one protocol and find parity rather than dominance: SAE-StatSteer matches the optimization-based SAE-SSV while exposing every decision for audit, and no single configuration passes our strict three-test, FDR-corrected criterion, so we emphasize effect sizes, uncertainty, clean success, and Pareto efficiency.
\squishend

\noindent\Cref{app:extended-related} connects this work to prior steering and evaluation research, and \cref{app:theory} develops the geometry behind each design choice.

\section{Related Work}
\label{sec:related}

Activation steering spans prompt-pair contrasts (ActAdd), dataset contrasts (CAA), head interventions \citep{li2023inferencetime}, and representation engineering (RepE); low-dimensional directions can causally mediate refusal \citep{arditi2024refusal}. Dense directions may entangle behavior with topic or style under superposition \citep{elhage2022superposition}, whereas SAEs provide sparse features \citep{cunningham2024sae} through JumpReLU and Gemma Scope dictionaries \citep{rajamanoharan2024jumprelu,deepmind2025gemmascope2}. Feature labels still require behavioral validation \citep{kantamneni2025sae,sharkey2025open}, and AxBench finds SAEs uncompetitive with simple steering baselines, so we compare against dense methods under one protocol.

SAE methods differ in selection and construction: SAE-RSV refines a dense direction with LLM-scored semantics; SAE-SSV optimizes within a supervised SAE subspace using a language-model loss; SAE-TS measures downstream rollout effects \citep{chalnev2024learning}; SAE-Steering combines keyword-logit recall with intervention ranking; and select-and-project methods aggregate top encoder features into a control axis and project it out \citep{barbalau2025rethinking}. SAE-StatSteer instead adds Cohen's-$d$-weighted decoder rows chosen by statistical consensus, without assuming that a row reproduces only its apparent concept. \Cref{tab:method_comparison} compares the closest methods.

LLM judges approximate preferences but exhibit position, verbosity, and related biases \citep{zheng2023judging,shi2024judging,li2024judgment,gu2024judgment}. We randomize order, aggregate three judges, and adjudicate disagreements by hand. Complementing the theoretical analysis of dense steering by \citet{im2025unified}, our protocol compares dense and SAE methods while separating raw movement from success conditioned on relevance, richness, and coherence.

\section{Problem Formulation}
\label{sec:problem}

Let $h^{(\ell)}(x)\in\mathbb{R}^{T\times d_{\text{model}}}$ be the residual stream at layer $\ell$. An SAE trained on that stream has decoder $W_{\text{dec}}\in\mathbb{R}^{D_{\textsc{sae}}\times d_{\text{model}}}$, whose rows are feature directions. For a behavioral attribute $A$, we constrain the steering vector to selected decoder rows:
\[
\delta\mathbf{h}=\sum_{i\in\mathcal{F}}w_i W_{\text{dec}}[i,:],
\]
where $\mathcal{F}$ is the selected feature set and $w_i$ its auditable weights. At inference, a strength $\alpha\geq0$ scales this direction at its source layer.

Let $g_0(x)$ and $g_\alpha(x)$ be baseline and steered generations, $s_A$ the target-attribute score, and $q$ a quality functional over relevance, richness, and coherence. We seek
\[
\begin{aligned}
\Delta_A &= \mathbb{E}_x\!\left[s_A(g_\alpha(x))-s_A(g_0(x))\right], \\
\max_{\ell,\mathcal{F},\alpha}\quad &\Delta_A
\quad \text{s.t.}\quad \mathbb{E}_x[q(g_\alpha(x))]\geq\tau .
\end{aligned}
\]
A useful configuration therefore lies on the Pareto frontier of attribute movement and quality preservation; raw movement alone is insufficient. Because the effective region varies across models and attributes, we search layer, feature set, and strength rather than fixing them globally. \Cref{sec:method} instantiates the direction, and \cref{sec:setup} defines the measured scores.

\section{Methodology}
\label{sec:method}

SAE-StatSteer has three auditable stages: contrastive feature extraction, consensus selection, and closed-form steering (\cref{fig:main_pipeline}).

\begin{table*}[t]
  \centering
  \resizebox{\textwidth}{!}{%
  \begin{tabular}{llcccccc}
    \toprule
    \textbf{Model} & \textbf{Domain} & \textbf{Pos.\ Configs} & \textbf{Mean $\Delta_p$} & \textbf{Best Raw} & \textbf{Best Quality-Pos.} & \textbf{Best Clean-Pos.} & \textbf{Robust Layer} \\
    \midrule
    \multirow{4}{*}{\gtwob}
      & \textsc{moral}     & 16/24 & $+0.037$ & L16, $\alpha{=}1.5$ ($+0.550$) & L12, $\alpha{=}1$ ($\Delta_q{=}{+}0.493$) & L19, $\alpha{=}0.1$ (11.0\%) & L12 (87.5\%) \\
      & \textsc{logic}     & 15/24 & $+0.121$ & L19, $\alpha{=}0.5$ ($+0.590$) & L19, $\alpha{=}1$ ($\Delta_q{=}{+}0.605$) & L19, $\alpha{=}0.7$ (17.0\%) & L19 (87.5\%) \\
      & \textsc{politics}  & 5/24  & $-0.146$ & L16, $\alpha{=}0.5$ ($+0.610$) & L16, $\alpha{=}1$ ($\Delta_q{=}{+}0.603$) & L16, $\alpha{=}1$ (25.0\%) & L16 (50.0\%) \\
      & \textsc{sentiment} & 5/16  & $-0.094$ & L13, $\alpha{=}0.2$ ($+0.400$) & L13, $\alpha{=}0.2$ ($\Delta_q{=}{+}0.463$) & L19, $\alpha{=}0.2$ (7.0\%) & L19 (50.0\%) \\
    \midrule
    \multirow{4}{*}{\gninb}
      & \textsc{moral}     & 20/32 & $+0.050$ & L38, $\alpha{=}0.5$ ($+0.700$) & L38, $\alpha{=}0.2$ ($\Delta_q{=}{+}0.527$) & L38, $\alpha{=}0.1$ (17.0\%) & L38 (87.5\%) \\
      & \textsc{logic}     & \cellcolor{robustPurpleMed}25/32 & \cellcolor{rawBlueLight}$+0.281$ & \cellcolor{rawBlueLight}\textbf{L19, $\boldsymbol{\alpha{=}0.1}$ ($\boldsymbol{+1.160}$)} & \cellcolor{qualMagentaLight}L19, $\alpha{=}0.5$ ($\boldsymbol{\Delta_q{=}{+}1.070}$) & L26, $\alpha{=}1$ (26.0\%) & \cellcolor{robustPurpleMed}L19 (\textbf{100.0\%}) \\
      & \textsc{politics}  & 15/32 & $-0.011$ & L38, $\alpha{=}1.5$ ($+0.612$) & L26, $\alpha{=}0.7$ ($\Delta_q{=}{+}0.398$) & \cellcolor{cleanPurpleLight}L19, $\alpha{=}0.5$ (30.6\%) & L31 (62.5\%) \\
      & \textsc{sentiment} & 22/32 & $+0.166$ & L26, $\alpha{=}0.3$ ($+0.640$) & L31, $\alpha{=}0.7$ ($\Delta_q{=}{+}0.407$) & L31, $\alpha{=}0.1$ (17.0\%) & L31 (87.5\%) \\
    \midrule
    \multirow{4}{*}{\gthrb}
      & \textsc{moral}     & 15/32 & $-0.012$ & L9, $\alpha{=}0.7$ ($+0.380$) & L9, $\alpha{=}0.3$ ($\Delta_q{=}{+}0.533$) & L9, $\alpha{=}0.7$ (10.0\%) & L22 (75.0\%) \\
      & \textsc{logic}     & 22/32 & $+0.140$ & L17, $\alpha{=}1$ ($+0.750$) & L22, $\alpha{=}0.5$ ($\Delta_q{=}{+}0.535$) & L17, $\alpha{=}1$ (26.0\%) & \cellcolor{robustPurpleMed}L22 (\textbf{100.0\%}) \\
      & \textsc{politics}  & 14/32 & $-0.051$ & \textbf{L22, $\boldsymbol{\alpha{=}2}$ ($\boldsymbol{+0.796}$)} & L22, $\alpha{=}2$ ($\Delta_q{=}{+}0.537$) & \cellcolor{cleanPurpleLight}L22, $\alpha{=}2$ (\textbf{34.7\%}) & L22 (62.5\%) \\
      & \textsc{sentiment} & 14/32 & $-0.026$ & L22, $\alpha{=}1$ ($+0.540$) & L9, $\alpha{=}1.5$ ($\Delta_q{=}{+}0.473$) & L22, $\alpha{=}0.3$ (10.0\%) & L9 (62.5\%) \\
    \bottomrule
  \end{tabular}}
  \caption{Headline single-layer results for three Gemma models and four behavioral domains. Columns give the number of configurations with positive mean primary delta, the mean $\Delta_p$ over a cell's configurations, and the best configuration by raw $\Delta_p$, by quality delta $\Delta_q$ among positive-shift configurations, and by clean success, plus the layer most often positive across strengths. Best entries are selected \textit{post hoc} on the evaluation prompts (\cref{sec:setup}). \colorbox{rawBlueLight}{\scriptsize Cyan} marks the largest mean $\Delta_p$ and raw shift ($+1.160$), \colorbox{qualMagentaLight}{\scriptsize pink} the largest quality-positive setting ($\Delta_q{=}{+}1.070$), \colorbox{cleanPurpleLight}{\scriptsize light lavender} the two clean-success rates above 30\%, and \colorbox{robustPurpleMed}{\scriptsize lavender} the layers positive at every strength and the largest positive count (25/32); \textbf{bold} marks the global maximum of each column. For \textsc{politics} and \textsc{sentiment}, positive deltas are rightward or positive polarity shifts, not quality improvements.}
  \label{tab:headline}
\end{table*}

\begin{figure*}[t]
    \centering
    \includegraphics[width=\linewidth]{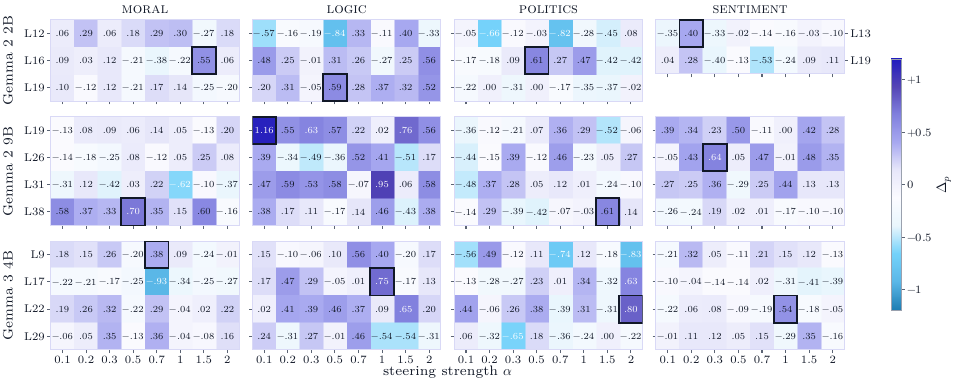}
    \caption{The layer--$\alpha$ steering landscape for three models (rows) and four domains (columns). Each cell gives the mean primary delta $\Delta_p$ (steered minus baseline) of one (layer, $\alpha$) configuration; purple marks positive shifts and cyan negative ones, with saturation tracking magnitude, and a frame marks each panel's best configuration. The operative layer is domain- and model-specific (\textsc{logic} peaks at L19 in Gemma~2 9B but L17 in Gemma~3 4B; \textsc{moral} at L16, L38, and L9), and the response to $\alpha$ is non-monotonic: several of the strongest cells occur at the smallest strength (Gemma~2 9B \textsc{logic}: $+1.16$ at $\alpha{=}0.1$).}
    \label{fig:heatmap}
\end{figure*}

\subsection{Contrastive Pairs and SAE Features}
\label{subsec:pairs}
\label{subsec:extraction}
For each domain $D\in\{\textsc{moral},\textsc{logic},\textsc{politics},\textsc{sentiment}\}$, we construct 800 topic-matched contrast pairs $\mathcal{P}_D=\{(x_j^+,x_j^-)\}_{j=1}^{800}$, which yields 1{,}600 balanced texts per domain. Topic matching reduces the chance that a selected feature reflects subject matter rather than the target contrast \citep{burns2023discovering}. Sources are IMDB \citep{maas2011imdb}, LogicBench \citep{parmar2024towards}, ETHICS \citep{hendrycks2021ethics}, and Twinviews-13k \citep{fulayRelationshipTruthPolitical2024}. Logic and politics use naturally contrasting material. Sentiment and moral retain benchmark anchors and use LLM-generated opposing counterparts, a possible source-style confound discussed in \cref{sec:limitations,app:methodology}.

We use width-16{,}384 JumpReLU SAEs from Gemma Scope for Gemma~2 and Gemma Scope~2 for Gemma~3 4B, all trained on the post-MLP residual stream. At candidate layer $\ell$, the SAE encodes $H^{(\ell)}(x_i)\in\mathbb{R}^{T\times d_{\text{model}}}$. We max-pool over tokens to obtain $z_i\in\mathbb{R}^{D_{\textsc{sae}}}$ because a sparse feature's peak activation retains localized evidence better than its sequence mean. Selection receives only $Z_D\in\mathbb{R}^{1600\times16384}$ and balanced labels $y_D$; \cref{app:models} lists models and layers.

\subsection{Three-Statistic Consensus Selection}
\label{subsec:selection}
We transform activations with $\operatorname{log1p}$ and rank every feature by three complementary criteria. The $F$-statistic detects class-conditional mean separation and supports closed-form significance testing. Cohen's $d$ supplies signed, variance-normalized separation and later determines each decoder row's weight. KSG mutual information captures nonlinear, multimodal, or threshold dependence that mean-based measures can miss. The statistics are classical; the design choice is their consensus. \Cref{tab:stat-comparison-main} summarizes their assumptions and failure modes, and \cref{app:stat-analysis} gives the estimators.

A feature must pass six reliability conditions: activity rate at least 1\% with at least five active rows; $d>0$; a positive lower 95\% bootstrap bound for $d$; $|d|\geq0.2$; MI bootstrap $\textsc{cv}<0.5$; and FDR-corrected $q<0.05$ for at least one test.\footnote{The second and fourth conditions together are equivalent to $d\geq0.2$. We keep them separate because $d>0$ is the directional switch (reversed for the inversion discussed in the Ethical Considerations), whereas $|d|\geq0.2$ is a sign-free floor whose value we stress-test in \cref{app:filter}.} The final condition is an OR filter, not a global FDR guarantee over the union: each test is corrected separately, and consensus then rewards shared evidence. \Cref{app:filter} motivates each threshold.

Surviving features receive ranks by $F$, debiased MI, and $|d|$. A high-confidence tier contains features in the top 300 for all three rankings; a medium-confidence tier contains those in the top 300 for exactly two. Within each tier, mean rank gives an unweighted Borda order \citep{emerson2013original}. Selection writes candidate sets of $K\in\{16,24,32\}$ features, filled from the high tier and then the medium tier, and keeps the whole consensus pool when the two tiers hold fewer than $K$ features. Each steered layer uses one set: $K{=}32$ for 31 of the 43 steered layer--domain pairs, $K{=}16$ for two, and the whole pool (3 to 31 features) for the other ten (\cref{tab:models}). This robustness-oriented rule may sacrifice sensitivity, but it resists artifacts from any single statistic. In our data, $F$ and $|d|$ are strongly rank-correlated ($\rho\in[0.87,0.99]$), whereas MI contributes the more independent signal ($\rho\in[0.27,0.80]$ with $F$).

\subsection{Steering Direction and Injection}
\label{subsec:vector}
Let $\mathbf{f}=(f_1,\ldots,f_K)$ be the selected indices and $\mathbf{d}$ their Cohen's-$d$ values. We form
\[
\mathbf{v}_{\text{init}} = \frac{\mathbf{d}}{\|\mathbf{d}\|_2},\qquad
\delta\mathbf{h} = \mathbf{v}_{\text{init}}^\top W_{\text{dec}}[\mathbf{f},:],
\]
and $\widehat{\delta\mathbf{h}}=\delta\mathbf{h}/\|\delta\mathbf{h}\|_2$. Cohen's $d$ is a signed standardized separation, so stronger and more stable class contrasts receive greater weight. Under diagonal, comparable within-class variances, this weighting also aligns with a Fisher LDA direction, a connection we use as motivation rather than an optimality claim (\cref{subsec:geometry}). The construction is closed form and introduces no learning rate, step count, or auxiliary loss.

\label{subsec:inference}
At the SAE source layer, a forward hook modifies the last-token residual state:
\[
h^{(\ell)}_{:,-1,:}\leftarrow h^{(\ell)}_{:,-1,:}+\alpha\,\|h^{(\ell)}_{:,-1,:}\|_2\,\widehat{\delta\mathbf{h}}.
\]
Activation-proportional scaling gives $\alpha$ the same geometric meaning in every configuration, and same-layer injection respects the activation distribution on which the SAE decoder was trained. We sweep $\alpha\in\{0,0.1,0.2,0.3,0.5,0.7,1.0,1.5,2.0\}$. The shared $\alpha{=}0$ run is the baseline and is excluded from the 344 nonzero-strength single-layer configurations. Multi-layer composition allocates a bounded global strength across selected layers (\cref{app:multilayer}).

\subsection{What the Geometry Guarantees}
\label{subsec:geometry}
Three properties, proved in \cref{app:theory} and drawn in \cref{fig:theory-direction,fig:theory-fisher,fig:theory-consensus}, make the construction predictable. First, C2 keeps every weight positive, so $\delta\mathbf{h}$ lies in the convex cone of the selected decoder rows and no selected feature pulls against another (\cref{prop:cone}). Second, Cohen's-$d$ weighting is the midpoint of the family $\mathbf{w}(p)=\Delta\boldsymbol{\mu}\odot\boldsymbol{\sigma}^{-p}$, which runs from the feature-space mean difference ($p{=}0$) to diagonal Fisher LDA ($p{=}2$); a Kantorovich bound keeps both ends within $\arccos\big(2\sqrt{\kappa}/(1{+}\kappa)\big)$ of it, where $\kappa$ is the ratio of the largest to the smallest standard deviation among the selected features (\cref{prop:kantorovich}). In the 41 filter-consistent released directions, the cosine to diagonal Fisher is 0.88 to 0.99, but to full-covariance Fisher only a median 0.68, because the selected features are correlated within class (\cref{fig:theory-fisher}). Third, activation-proportional injection rotates the last-token state by an angle that depends on $\alpha$ alone, $\arctan\alpha$ when $\widehat{\delta\mathbf{h}}\perp h$ (\cref{prop:injection}): $45^\circ$ at $\alpha{=}1$ and $63^\circ$ at the ceiling $\alpha{=}2$, whatever the scale of the layer.

\begin{table*}[t]
    \centering
    \small
    \setlength{\tabcolsep}{3.4pt}
    \renewcommand{\arraystretch}{1.06}
    \begin{tabular}{l cccc cccc cccc c c}
        \toprule
        & \multicolumn{4}{c}{\textbf{\gtwob}}
        & \multicolumn{4}{c}{\textbf{\gninb}}
        & \multicolumn{4}{c}{\textbf{\gthrb}} & & \\
        \cmidrule(lr){2-5}\cmidrule(lr){6-9}\cmidrule(lr){10-13}
        \textbf{Method}
        & \textbf{Sent.} & \textbf{Pol.} & \textbf{Mor.} & \textbf{Log.}
        & \textbf{Sent.} & \textbf{Pol.} & \textbf{Mor.} & \textbf{Log.}
        & \textbf{Sent.} & \textbf{Pol.} & \textbf{Mor.} & \textbf{Log.} &
        & \textbf{Mean} \\
        \midrule
        CAA          & 34.5 & 32.1 & 28.4 & 24.7 & 42.6 & 30.1 & 33.0 & 27.6 & 35.5 & 43.2 & 29.6 & 23.1 && 32.0 \\
        RePE         & 22.2 & 30.0 & 20.7 & 17.1 & 17.8 & 20.6 & 18.6 & 14.0 & 21.0 & 32.9 & 20.1 & 16.0 && 20.9 \\
        Top-PC       & 20.7 & 28.5 & 18.0 & 16.0 & 18.5 & 24.2 & 17.5 & 15.2 & 19.0 & 30.6 & 18.0 & 15.7 && 20.2 \\
        ITI          & 35.1 & 39.0 & 27.0 & 25.3 & 41.4 & 26.5 & 20.1 & 28.5 & 34.1 & 42.0 & 28.3 & 24.3 && 31.0 \\
        SAE-SSV      & \textbf{38.1} & 29.1 & 32.3 & 27.8 & \textbf{47.5} & \textbf{39.2} & \textbf{45.6} & 33.1 & \textbf{42.6} & 45.1 & 34.7 & 27.0 && 36.8 \\
        SAE-RSV      & 27.8 & 37.7 & 30.8 & 25.0 & 42.6 & 35.7 & 42.1 & 30.2 & 38.0 & 42.8 & 29.5 & 25.8 && 34.0 \\
        SAE-Steering & 31.1 & 35.0 & 28.4 & 21.6 & 37.6 & 34.1 & 39.0 & 27.8 & 31.1 & 40.4 & 32.0 & 25.1 && 31.9 \\
        \midrule
        \rowcolor{lavtint}
        SAE-StatSteer (\textit{post hoc}) & 32.2 & \textbf{40.4} & \textbf{36.5} & \textbf{33.6} & 38.2 & 37.8 & 45.4 & \textbf{36.6} & 33.8 & \textbf{48.0} & \textbf{37.6} & \textbf{31.0} && \textbf{37.6} \\
        \bottomrule
    \end{tabular}
    \caption{Raw success rate (SR, \%) of all seven baselines and SAE-StatSteer per model and domain under one three-judge protocol (\cref{sec:setup}), with the mean over the 12 cells; \textbf{bold} marks the highest SR per column. Baselines run at their recommended configurations, whereas SAE-StatSteer reports each cell's best configuration, chosen \textit{post hoc} on the evaluation prompts. SAE-StatSteer has the highest SR in 7 of 12 cells and SAE-SSV in the other 5, including all three \textsc{sentiment} cells; their means differ by 0.8 points, which we read as parity rather than dominance.}
    \label{tab:ablation-study}
    \label{tab:baselines}
\end{table*}

\begin{table*}[t]
  \centering
  \small
  \setlength{\tabcolsep}{2.2pt}
  \resizebox{\textwidth}{!}{%
  \begin{tabular}{lcccc}
    \toprule
    \textbf{Property}
      & \makecell[c]{\textbf{SAE-SSV}\\[-1pt]{\footnotesize\citep{he2025saessv}}}
      & \makecell[c]{\textbf{SAE-RSV}\\[-1pt]{\footnotesize\citep{wang2025saersv}}}
      & \makecell[c]{\textbf{SAE-Steering}\\[-1pt]{\footnotesize\citep{fang2026saesteering}}}
      & \textbf{Ours} \\
    \midrule
    Feature selection signal              & Linear probe          & LLM semantic judge & Causal probing    & 3-statistic Borda \\
    No learned aggregation weights        & \xmark                & \xmark             & \xmark            & \cellcolor{cleanPurpleLight}\cmark \\
    Optimization needed                   & \cmark                & \xmark             & \cmark            & \cellcolor{cleanPurpleLight}\xmark \\
    Per-statistic FDR correction          & \xmark                & \xmark             & \xmark            & \cellcolor{cleanPurpleLight}\cmark \\
    Quality-conditioned evaluation        & Partial               & \xmark             & \xmark            & \cellcolor{cleanPurpleLight}\cmark \\
    Per-feature interpretability          & Partial               & \cmark             & \cmark            & \cmark \\
    Models evaluated                      & Llama, Qwen, Gemma    & Llama-3-8B         & DeepSeek-R1, Qwen & \glogo Gemma 2/3 \\
    Behavioral domains                    & 3                     & 5                  & 5 (reasoning)     & 4 \\
    Layer--$\alpha$ configurations surveyed per model & 1 layer & 1 layer          & 1 layer           & \cellcolor{robustPurpleMed}\textbf{88 / 128 / 128} \\
    \bottomrule
  \end{tabular}}
  \caption{Design-level comparison with the closest SAE-steering baselines. \colorbox{cleanPurpleLight}{\scriptsize Light lavender} marks properties that our pipeline provides under the tested setting, and \colorbox{robustPurpleMed}{\scriptsize lavender} the larger layer--strength sweep, counted per model (344 in total). ``Quality-conditioned evaluation'' is \textit{Partial} for SAE-SSV, which reports disorder and coherence diagnostics; our contribution is the joint clean-success criterion over relevance, richness, and coherence.}
  \label{tab:method_comparison}
\end{table*}

\section{Experimental Setup}
\label{sec:setup}

\paragraph{Models and data.} We study Gemma~2 2B, Gemma~2 9B \citep{gemmateam2024gemma2}, and Gemma~3 4B \citep{gemmateam2025gemma3}, probing early, middle, and late layers (\cref{app:models}). Feature selection uses 800 contrast pairs per domain, for 3{,}200 pairs in total. Evaluation uses 100 held-out prompts per domain, disjoint from selection: LogicBench for \textsc{logic}, Twinviews-13k for \textsc{politics}, and synthetic incomplete sentences for \textsc{moral} and \textsc{sentiment}. Generation uses temperature 0.2 and at most 200 new tokens.

\paragraph{Operating points and search budget.} The grid was fixed \textit{a priori}. Candidate layers are early, middle, and late layers of each model, following prior SAE-steering practice, and only layers that elicited a maximal response for some domain were steered (Gemma~2 2B L23 never did). The strength ceiling $\alpha{=}2$ was set on a disjoint validation set of 50 prompts per domain, above which outputs degenerated. Each steered layer uses one $K$, so the sweep has 344 nonzero-strength configurations (88, 128, and 128 per model) rather than 1{,}032. The best-per-cell entries of \cref{tab:headline,tab:baselines} are therefore \textit{post hoc} choices on the 100 evaluation prompts, and we label them as such. Baselines run at their recommended configurations; matching our sweep for all seven would need about 241k further judged generations, so search budgets are unmatched (\cref{app:budgets}).

\paragraph{Baselines.} We compare four dense methods, Contrastive Activation Addition (CAA), Representation Engineering (RePE), top-principal-component steering (Top-PC), and Inference-Time Intervention (ITI), with three SAE methods: SAE-SSV, SAE-RSV, and SAE-Steering. Every method uses the same three models, held-out prompts, and evaluation protocol. \Cref{app:baseline-details} documents intervention layers, SAE dictionaries, implementation provenance, and the out-of-design adaptation of SAE-Steering.

\paragraph{Judging protocol.} Gemini~2.5 Flash, Gemini~2.5 Pro \citep{comanici2025gemini25}, and GPT-5.4 \citep{openai2026gpt54} independently compare each steered output with its baseline, with A/B order randomized to limit position bias. Each judge assigns 1--10 scores for the domain-specific primary attribute, relevance, richness, coherence, and factuality, and returns a pairwise winner. When all three verdicts agree, the verdict is accepted and the scores are averaged over judges. The remaining split items (at most 2\%) go to an author who sees the prompt, both outputs, and every judge's scores and explanations, and whose verdict is final. We also checked 78 sampled unanimous items by hand; \cref{app:eval} gives the rubric and the prompt schema.

\paragraph{Metrics and interpretation.} Primary delta is $\Delta_p=\bar p_{\text{steered}}-\bar p_{\text{baseline}}$; quality delta $\Delta_q$ averages the relevance, richness, and coherence changes. Factuality is scored but left out of $\Delta_q$ by design, because the \textsc{logic} primary score already measures ground-truth correctness and the other domains elicit narrative or opinion. Success rate (SR) is the fraction of prompts whose adjudicated steered primary score exceeds the baseline's. Clean success additionally requires relevance${\geq}7$, richness${\geq}4$, and coherence${\geq}4$. For \textsc{moral} and \textsc{logic}, a positive primary change means more ethical or more correct output, so these are improvement-oriented domains. For \textsc{politics} and \textsc{sentiment}, it means a rightward or more positive polarity shift, not a normative improvement, so we interpret them as directional-control domains. Two \textsc{politics} items lack valid judgments in each of the 64 \textsc{politics} configurations of Gemma~2 9B and Gemma~3 4B; those cells use $n{=}98$, the analysis holds 34{,}272 of 34{,}400 paired rows, and the missing rows are excluded rather than imputed.

\paragraph{Release.} The linked repository holds all contrast pairs, prompts, steering vectors, and code.

\section{Results}
\label{sec:results}

Across the sweep, two findings recur: effective steering is localized by model, domain, layer, and strength, and raw attribute movement overstates quality-preserving control. \Cref{tab:headline} summarizes all 12 model--domain cells, and \cref{fig:heatmap} shows the 344 nonzero-strength single-layer settings. Best cells are descriptive Pareto candidates; conservative configuration-level tests follow in \cref{subsec:stat-rigor}.

\subsection{Single-Layer Landscape}
\label{subsec:improve}
\label{subsec:directional}
\label{subsec:non_universal}
Logic is the strongest improvement-oriented domain. Gemma~2 9B reaches $\Delta_p{=}{+}1.160$ at L19 and $\alpha{=}0.1$, the largest shift in the sweep; Gemma~3 4B and Gemma~2 2B peak at $+0.750$ and $+0.590$. Moral peaks are $+0.700$, $+0.550$, and $+0.380$, but clean success rarely exceeds 17\%. Directional-control peaks are $+0.796$ for Gemma~3 4B politics, with 34.7\% clean success, and $+0.640$ for Gemma~2 9B sentiment.

The operative setting is not universal. Logic peaks at L19 in Gemma~2 9B but L17 in Gemma~3 4B; moral peaks at L16, L38, and L9 across the three models. Relative depth has nearly zero rank correlation with primary delta ($\rho{=}{-}0.005$, $p{=}0.98$, $N{=}344$), as does $\alpha$ ($\rho{=}{-}0.017$, $p{=}0.82$). \Cref{fig:heatmap} also shows non-monotonic responses, so both layer and strength require model--domain selection. By model, Gemma~2 9B has a positive $\Delta_p$ in 64.1\% of its configurations and the largest mean shift ($+0.121$), whereas Gemma~2 2B has the largest mean $\Delta_q$ ($+0.081$); \cref{app:top_configs,app:extended-results} list these aggregates and the top configurations.

\begin{figure*}[t]
    \centering
    \includegraphics[width=\textwidth]{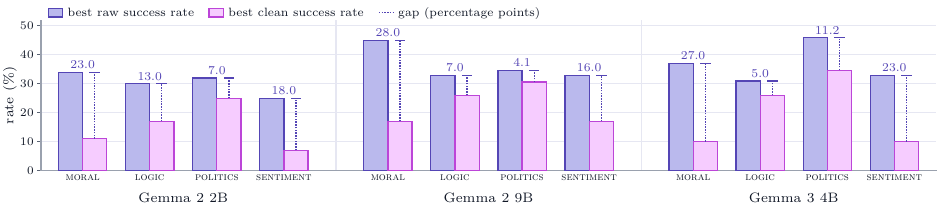}
    \caption{Best raw success rate and best clean success rate in each model--domain cell; the dotted bracket and the number above each pair give the gap in percentage points. Clean success additionally requires relevance${\geq}7$, richness${\geq}4$, and coherence${\geq}4$, so it can never exceed raw success (\cref{app:theory}). The gap ranges from 4.1 points (Gemma~2 9B \textsc{politics}) to 28.0 points (Gemma~2 9B \textsc{moral}).}
    \label{fig:gap}
\end{figure*}

\begin{table}[t]
  \centering
  \small
  \setlength{\tabcolsep}{4pt}
  \begin{tabular}{l>{\raggedright\arraybackslash}p{0.70\columnwidth}}
    \toprule
    \textbf{Domain} & \textbf{Top failure signals (pooled over models)} \\
    \midrule
    \textsc{moral}     & repetitive (57.4\%), repetition (51.8\%), coherence (44.1\%), incoherent (17.5\%), unethical (8.5\%) \\
    \textsc{logic}     & incorrect (53.3\%), repetitive (29.9\%), incoherent (18.1\%), contradicts (12.6\%), irrelevant (10.6\%) \\
    \textsc{politics}  & repetitive (41.0\%), coherence (29.0\%), incoherent (24.7\%), repetition (21.8\%), irrelevant (14.7\%) \\
    \textsc{sentiment} & repetition (52.3\%), repetitive (47.8\%), coherence (41.3\%), off-topic (7.7\%), incoherent (6.3\%) \\
    \bottomrule
  \end{tabular}
  \caption{Dominant judge-flagged failure signals by domain, pooled over the three models; flags are nonexclusive keyword categories from the judges' explanations. Repetition and incoherence recur in every domain, which motivates clean success as the principal metric.}
  \label{tab:failures}
\end{table}

\subsection{Raw Movement versus Usable Control}
\label{subsec:failure}
Best raw SR exceeds best clean success by 4.1 to 28.0 percentage points per cell (\cref{fig:gap}). \Cref{tab:failures} identifies the recurring failure signals behind this gap, which \cref{subsec:qualitative-audit} connects to selected-feature semantics and sample-level transitions. The gap therefore often reflects degraded output after intervention, not failure to move the target attribute. Clean success is raw success restricted to a region of quality space, so the gap is the share of raw successes that leave it (\cref{rem:clean}). The recurring flags name the face through which an output leaves: repetition lowers richness, incoherence lowers coherence, and off-topic drift lowers relevance (\cref{fig:theory-clean}).

\subsection{Which Components Matter}
\label{subsec:ablation}

\begin{table}[t]
  \centering
  \small
  \setlength{\tabcolsep}{5pt}
  \begin{tabular}{lcc}
    \toprule
    \textbf{Variant} & $\Delta_p\uparrow$ & \textbf{Clean SR (\%)}$\uparrow$ \\
    \midrule
    \rowcolor{lavtint}
    Full pipeline              & \textbf{0.064} & \textbf{19.28} \\
    $F$-only ranking           & 0.012 & 2.70 \\
    MI-only ranking            & 0.017 & 1.51 \\
    Cohen's-$d$-only ranking   & 0.038 & 9.60 \\
    No reliability filter      & 0.014 & 5.80 \\
    \bottomrule
  \end{tabular}
  \caption{Component ablation, pooled over settings under the same protocol. Single-statistic rankings replace the Borda consensus; the filter-free variant ranks all active features. Consensus doubles clean success over the best single statistic, and removing the filter cuts it by 70\%.}
  \label{tab:ablation}
\end{table}

\Cref{tab:ablation} isolates the pipeline's parts. Pooled over settings under the same protocol, the full pipeline reaches $\Delta_p{=}0.064$ and 19.28\% clean success. Ranking by any single statistic loses at least half of the clean success (9.60\% at best, for Cohen's $d$), and dropping the reliability filter loses 70\% of it (5.80\%). Although $F$ and $|d|$ are strongly rank-correlated, $F$-only and $d$-only ranking differ threefold in $\Delta_p$, so the consensus is not redundant. On weighting, the released directions explain why it is second-order: within a direction, the largest selected $d$ is only a median 2.7 times the smallest, so $d$-weighted and uniformly weighted residual directions have a median cosine of 0.978 (\cref{app:theory}).

\subsection{Multi-Layer Steering}
\label{subsec:multilayer}
Bounded multi-layer composition does not amplify the best raw single-layer shift: its raw $\Delta_p$ is lower in all 12 cells. The injection geometry is consistent with this: every hook receives $\alpha_\ell\leq1$, which caps its rotation at $45^\circ$ in the orthogonal case, against up to $63^\circ$ for the strongest single-layer points (\cref{prop:injection}). It can, however, improve reliability. Gemma~2 9B logic rises from 24\% clean success at the raw-max single-layer point and 26\% at the best-clean point to 31\%, gains of 7 and 5 points. Gemma~2 2B logic gains 4 points over its raw-max reference but remains 1 point below its 17\% best-clean rate (\cref{tab:multilayer-main}).

\subsection{Comparison to Prior Steering Methods}
\label{subsec:baseline-comparison}
\Cref{tab:baselines} gives the head-to-head comparison under the common judging protocol, and \cref{tab:method_comparison} summarizes design and implementation differences. With \textit{post hoc} operating points, SAE-StatSteer has the highest raw SR in 7 of the 12 model--domain cells, including all three logic cells, and SAE-SSV in the other five, including sentiment for all three models; on Gemma~2 9B moral the two differ by 0.2 points (45.4 \vs 45.6). Averaged over cells, their raw SR is 37.6\% and 36.8\%. We therefore claim parity with the optimization-based SAE-SSV, without gradient optimization and with every decision exposed, rather than uniform dominance: the comparison is domain-specific, and a trained subspace captures the smooth polarity axis of sentiment better.

\subsection{Robustness}
\label{subsec:stat-rigor}
Across all 344 single-layer configurations, primary delta correlates moderately with quality delta (Spearman $\rho{=}0.463$, $p{<}10^{-17}$) and primary-win rate ($\rho{=}0.442$), and negatively with bad-output rate ($\rho{=}{-}0.301$). Under the strict joint requirement that FDR-corrected paired $t$-, Wilcoxon, and sign tests all have $p<0.05$, no individual configuration survives. The scores are discrete and tie-heavy, with a median tie rate near 44\%, and correction spans all configurations. The strongest cells still have intervals clear of zero: Gemma~2 9B \textsc{logic} at L19, $\alpha{=}0.1$ has a bootstrap 95\% CI of $[+0.33,+1.96]$ (Cohen's $d_z{=}0.276$), and Gemma~2 9B \textsc{moral} at L38, $\alpha{=}0.5$ one of $[+0.28,+1.15]$ ($d_z{=}0.311$). The first is also Pareto-efficient on ($\Delta_p$, $\Delta_q$, clean rate), as is Gemma~3 4B \textsc{politics} at L22, $\alpha{=}2$, the only efficient setting in its cell and the one with the highest clean success (\cref{tab:stat-pareto}). We therefore rely on effect sizes, bootstrap intervals, clean success, and Pareto efficiency rather than isolated significance claims; \cref{app:stat-holistic} gives the full tests.

\subsection{Feature- and Output-Level Audit}
\label{subsec:qualitative-audit}

\paragraph{Selected-feature semantics.}
For Gemma~2 2B layer~12 \textsc{logic}, the selected IDs map coherently to mathematical and logical expressions (13079); conditionals (6967, 10680); causal relations and dependencies (14319, 5560); reasoning transitions (11771); negation and affirmation (12456); quantifiers and comparisons (26); definitions and implications (12731); and rule interpretation (13752). This is descriptive evidence of an inspectable logic-related subspace, not causal proof; \cref{tab:feature_descriptions} gives the full explanations.

\paragraph{Failure diagnostics.}
Among the nonexclusive judge flags (\cref{tab:failures}), repetition leads in \textsc{moral} (57.4\%), \textsc{politics} (41.0\%), and \textsc{sentiment} (52.3\%), followed by coherence-related flags, and incorrectness leads in \textsc{logic} (53.3\%), followed by repetition and incoherence. Their recurrence shows why target movement can overstate usable control.

\paragraph{Representative transitions.}
Gemma~3 4B \textsc{logic} moves from 1 to 10: the steered response matches the ground-truth ``no'' and explains why the premises do not entail that apples are absent. Gemma~2 9B \textsc{moral} moves from 2 to 7, becoming safe and coherent while retaining duplicated phrasing. These cases expose both substantive change and the residual defects that clean success captures.

\section{Discussion}
\label{sec:discussion}

The sweep disabuses us of the idea of a canonical steering layer or strength. Even within one architecture family, different attributes peak at different depths, and neither relative depth nor $\alpha$ has a monotone global association with primary movement. This does not establish a mechanistic hierarchy in which lexical and abstract attributes occupy fixed depths. It establishes a practical requirement: an activation-steering method should search and report its operating point for each model and domain rather than inherit one layer from prior work. Reporting only the best result of an undocumented search would hide this instability; the landscape and the selection rule are part of the result.

The raw--clean gap changes how steering success should be measured. A perturbation may move the target score while making the generation repetitive, incoherent, or irrelevant, and such output is not useful behavioral control even if its primary score improves. Quality conditioning therefore belongs in the success definition, not only in a secondary fluency diagnostic, and the gap is a stringent litmus test that raw success alone cannot pass. The same lesson qualifies claims built on interpretable SAE labels: an auditable feature basis explains what was injected, but it does not guarantee a usable generation, so behavioral and quality validation remain necessary.

SAE-StatSteer's closed-form direction removes optimization choices while reaching parity with learned methods, most clearly on logic and moral steering. Its consistent sentiment deficit against SAE-SSV marks the boundary of that advantage: a trained subspace can better capture some smooth polarity axes. The ablation is salutary here: with the geometry of the released directions, it locates the transparency dividend in selection rather than weighting, and because the reliability filter needs only activations and labels, it could precede other SAE methods, for instance SAE-SSV's $F$-ranked probe stage. Multi-layer composition has a similarly specific role: it trades raw amplification for clean success. These findings favor transparent selection when auditability and robust improvement are central, learned optimization when a domain clearly benefits from it, and multi-layer injection when quality matters more than raw amplification. All of them concern the Gemma family; whether they hold for other families is an open question that our data do not address.

\section{Conclusion}
\label{sec:conclusion}

We introduced SAE-StatSteer, an optimization-free activation-steering pipeline that selects SAE features by a Borda consensus of an $F$-test, KSG mutual information, and Cohen's $d$, then combines decoder rows with signed Cohen's-$d$ weights. Selection carries most of its clean success: dropping the consensus ranking or the reliability filter at least halves it. Across three Gemma-family models, four domains, 344 single-layer settings, 12 multi-layer mixtures, and seven baselines, two findings recur within this scope. Raw success can exceed quality-conditioned clean success by up to 28 percentage points, and effective layer and strength are model- and domain-specific rather than universal. SAE-StatSteer reaches parity with the optimization-based SAE-SSV, ahead on logic and moral behavior and behind on sentiment, so it offers a trade-off between transparency and performance rather than a panacea. Future work should select operating points on held-out prompts, compare closed-form and gradient-refined directions within the same feature subspace, develop principled multi-layer budgets, and test compositional steering across attributes and model families. Reliable steering is not movement alone, but intended change with capabilities intact.

\clearpage
\section{Limitations}
\label{sec:limitations}

\paragraph{Causal validation is statistical, not mechanistic.} Our feature-selection procedure ranks SAE features by statistical association with the class label; on its own, it does not establish that activating these features at inference causes the observed behavioral shift. The $\alpha$ sweep partially closes this loop, but a definitive causal account would require per-feature ablation or patching experiments, which we leave to future work.

\paragraph{We validate by outcome, not by downstream feature effect.} A decoder row's apparent semantic label does not guarantee that steering with it activates only that concept: a feature associated with one concept can produce unrelated downstream effects once used for steering \citep{chalnev2024learning}. Our reliability filtering and quality-conditioned evaluation are designed to catch the symptom of this failure mode, degraded or off-target generation, but we do not measure which downstream SAE features actually activate in steered rollouts. The selected rows also overlap: projecting a released direction back onto its rows misstates the weights by a median 51\% (\cref{app:theory}). A feature-effect analysis in the style of \citet{chalnev2024learning} would give a more direct mechanistic account.

\paragraph{Selection reuse and unmatched tuning budgets.} Our best-per-cell operating points are chosen on the same 100 evaluation prompts that score them, so SAE-StatSteer's entries in \cref{tab:baselines} are optimistic relative to baselines that run at one recommended configuration. Selecting the operating point on held-out prompts and giving every baseline a matched layer and strength sweep (about 241k further judged generations) would make the comparison cleaner.

\paragraph{Multi-judge dependence and adjudication scope.} Although we use three frontier judges with human adjudication on disagreement, adjudication touches only the split subset. Systematic miscalibration shared by all three judges, if any, would propagate into the agreed verdicts untouched; our manual check of 78 unanimous items (\cref{app:eval}) samples this risk but cannot exclude it.

\paragraph{Tied ordinal scores and missing outcomes constrain testing.} Judge scores are 1--10 ordinals with a non-trivial tie rate per cell, and 100 evaluation prompts per domain is a modest sample for detecting effects of an individual configuration. This motivates the conservative framework of \cref{subsec:stat-rigor}, which applies a strict joint criterion with FDR correction across all 344 configurations. Finer-grained scoring (\eg a continuous 0--100 scale) or more prompts per configuration would sharpen these tests. In addition, 128 of the 34{,}400 paired rows (two \textsc{politics} items in each of 64 configurations) have no valid judgment and are excluded.

\paragraph{Synthetic counter-sides in two of four domains.} Contrastive pairs for \textsc{sentiment} and \textsc{moral} pair a benchmark anchor with an LLM-generated opposing counter-side, whereas \textsc{logic} and \textsc{politics} use naturally contrastive material. The shared topic anchor mitigates topic-level confounding, but a selected feature in these two domains could still track human-written versus model-generated style rather than the target attribute alone. \textsc{logic} and \textsc{politics}, source-matched by construction, show the same localization and raw--clean gap, so we treat the confound as reduced rather than eliminated.

\paragraph{One-dimensional political axis.} We score politics on a single left--right scale. Real political positions vary along several dimensions, so a rightward shift on this axis summarizes, and may hide, changes along others.

\paragraph{Limited model coverage.} Our study covers three Gemma-family models, and every finding is scoped to that family. The cross-model patterns we report, including the strongest \textsc{logic} and \textsc{moral} effects at different depths, may not transfer to other families such as Llama, Mistral, or Qwen.

\paragraph{English-only evaluation.} All contrast pairs and generations are in English. Cross-lingual extension would require both translated pairs and SAEs trained on multilingual residual-stream data.

\paragraph{Single domain per steering vector.} Each steering vector targets a single behavioral axis. Compositional steering that simultaneously shifts, for example, sentiment and political polarity introduces interaction effects we have not characterized; naive vector addition is unlikely to suffice.

\section{Ethical Considerations}
\label{sec:ethics}

Activation steering is a dual-use technique. The same machinery that can shift a model toward more ethical, more correct, or more factual generation can equally shift it toward the opposite. Our \textsc{politics} and \textsc{sentiment} domains are framed throughout as directional-control experiments precisely to avoid the implication that ``rightward'' or ``more positive'' is normatively better; we report these results as evidence of what steering can do, not as endorsements of any direction. We urge practitioners to (i)~report directional metrics alongside quality metrics to discourage attribute-shift-only success claims; (ii)~accompany any deployed steering intervention with documentation of the contrast pairs, SAE checkpoint and layer, and $\alpha$ value, so that the intervention remains auditable; and (iii)~refrain from using activation steering as a substitute for fine-tuning where the behavior has high-stakes downstream consequences (medical or legal decision-making, for example) without additional human review of steered outputs.

A second concern is misuse of the \textsc{moral} domain. Our pipeline can, in principle, be inverted: by selecting features with $d<0$ instead of $d>0$, one could construct a vector that steers the model toward unethical content. We deliberately avoid documenting that inversion in detail and do not release inverted vectors, even though the symmetry is mathematically immediate. The SAEs themselves are public, and the contrastive datasets either derive from public benchmarks or are constructed by the procedures described here; we believe the marginal uplift to a malicious actor from our specific pipeline is small relative to the methodological transparency it offers to safety researchers.

\section*{Acknowledgments}

We convey our heartfelt gratitude to the anonymous reviewers for their constructive criticisms and insightful feedback, which were conducive to the improvement of the research work outlined in this paper. We also appreciate the Systems and Software Lab (SSL) of the Islamic University of Technology (IUT) for the generous provision of computing resources during the course of this project. Syed Rifat Raiyan, in particular, wants to thank his parents, Syed Sirajul Islam and Kazi Shahana Begum, for everything.

\bibliography{ref_CR}

\clearpage
\appendix
\raggedbottom
\section*{Appendix}

\begin{tcolorbox}[enhanced, colback=boxback, colframe=boxframe, boxrule=0.5pt, arc=2pt, left=5pt, right=5pt,
  top=4pt, bottom=2pt, fontupper=\small]
\appentry{app:methodology}{Method Details}{sample size, pair provenance, the six filter thresholds, the choice of $K$, the two roles of Cohen's $d$, and the hook implementation}
\appentry{app:theory}{Theory: The Geometry of SAE-StatSteer}{the decoder cone, the Cohen--Fisher family and its Kantorovich bound, injection geometry, consensus in rank space, and the clean region, with 3D figures}
\appentry{app:stat-analysis}{Feature-Level Statistics}{preprocessing, the $F$-test, KSG mutual information, Cohen's $d$, and their blind spots}
\appentry{app:domains}{Domain Specifications}{targets, anti-targets, and the sign convention of each domain}
\appentry{app:models}{Models, SAEs, and Steered Layers}{dictionaries, candidate layers, steered layers, and $K$ per layer}
\appentry{app:budgets}{Operating Points, Budgets, and Missing Data}{per-method search budgets, bootstrap stability, and the 128 absent rows}
\appentry{app:eval}{Evaluation Protocol}{held-out prompts, the five-axis rubric, judge aggregation, adjudication, and our manual check}
\appentry{app:baseline-details}{Baselines}{implementation details and the out-of-design adaptation of SAE-Steering}
\appentry{app:multilayer}{Multi-Layer Steering}{layer selection, weights, and comparisons with single-layer steering}
\appentry{app:extended-results}{Extended Results}{model aggregates, domain detail, interpretation, top configurations, and the Pareto view}
\appentry{app:stat-holistic}{Configuration-Level Statistics}{the test suite, correlations, regressions, Pareto sets, and conservative testing}
\appentry{app:qualitative}{Qualitative Audit}{Neuronpedia explanations of selected features and example generations}
\appentry{app:extended-related}{Extended Related Work}{dense steering, SAEs and their limits, SAE-based steering, and evaluation}
\end{tcolorbox}

\section{Method Details}
\label{app:methodology}

This appendix expands the pipeline of \cref{sec:method}. The formal setup in \cref{sec:problem} constrains the steering direction to the span of decoder rows learned at the intervention layer, which preserves a feature-level audit trail: each selected index identifies both the evidence used during selection and the residual-space direction used at inference.

\subsection{Sample-Size Justification}
At $n_+{=}n_-{=}800$ (1{,}600 samples per binary comparison), the $F$-test has $(1,1598)$ degrees of freedom with critical $F\!\approx\!3.84$ at $\alpha{=}0.05$ uncorrected; after BH correction over 16{,}384 features, the effective threshold rises substantially while retaining power for $|d|\!\approx\!0.2$. KSG MI variance scales as $\mathcal{O}(1/N)$; bootstrap standard deviations are $5$--$10\%$ of the point estimate for informative features. For Cohen's $d\!\approx\!1.0$, the analytical SE is $\sqrt{2/800+1/3200}\approx 0.054$, yielding 95\% CIs of approximately $d\pm 0.10$.

\subsection{Pair-Generation Provenance}
\label{app:provenance}
Each domain's 800 pairs integrate two data sources to balance scalability with linguistic nuance.
\squishlist
    \item \textbf{Public-benchmark adaptation.} We adapt established datasets to seed our pairs: IMDB \citep{maas2011imdb} for \textsc{sentiment}, Twinviews-13k \citep{fulayRelationshipTruthPolitical2024} for \textsc{politics}, LogicBench \citep{parmar2024towards} for \textsc{logic}, and ETHICS \citep{hendrycks2021ethics} for \textsc{moral}. The 800 anchors of each domain are sampled at random from its source dataset.
    \item \textbf{LLM-generated counter-sides.} For \textsc{sentiment} and \textsc{moral}, we take the positive or ethical anchor texts from the source benchmarks and generate an opposing counter-side on the same topic for each with GPT-5.4 \citep{openai2026gpt54}. For \textsc{logic} and \textsc{politics}, contrasting pairs come directly from their datasets.
\squishend
The dataset uses a uniform pair convention across all four domains: $d > 0$ indicates that a feature ``fires more for the target attribute,'' so the selection pipeline applies a single global target-direction filter without per-domain sign flipping. All pairs, prompts, steering vectors, and code are in the repository linked on the first page.

\subsection{The Six Quality-Filter Conditions}
\label{app:filter}

The six conditions fall into two functional groups: C1, C3, and C5 ensure that the statistics are well-defined and reliable, while C2, C4, and C6 enforce the directional, practical, and statistical requirements a steering feature must meet.

\textbf{C1 (activity rate $\geq 1\%$, $\geq 5$ non-zero rows, $\mathrm{var}>10^{-12}$).} This is a precondition for every downstream statistic to be well-defined, not a modeling choice. SAE features are sparse, so most fire on too few examples to support a reliable estimate: the $F$-test's pooled-variance denominator and Cohen's $d$'s pooled standard deviation become unstable with fewer than five non-zero observations, and the KSG estimator needs enough non-zero samples to form meaningful neighborhoods. The $1\%$ floor ($\approx 16$ of $1600$) and the hard $\geq 5$ requirement guarantee that each test runs on sufficient data, while the variance floor drops constant features that would produce division by zero. The threshold is deliberately low so that sparse but informative features survive; typically $1{,}000$--$4{,}500$ of $16{,}384$ features pass, indicating that the filter removes ill-defined cases without being over-aggressive.

\textbf{C2 ($d>0$, target direction).} This encodes the steering goal. A feature with $d>0$ fires more strongly for the target class and one with $d<0$ for the anti-target; both carry signal, but only the former points where we wish to steer. Restricting to $d>0$ makes every weight in $\mathbf{v}_{\text{init}}=\mathbf{d}/\|\mathbf{d}\|$ positive, so the direction is a coherent push toward the target rather than a combination of opposing pulls (\cref{prop:cone} states this as cone membership), and a single global filter operates across all four domains without per-domain sign flipping.

\textbf{C3 (one-sided bootstrap CI lower bound $>0$).} C2 uses a point estimate, which can be positive by chance for a noisy feature; C3 upgrades it to a reliability requirement by demanding that the lower bound of the $95\%$ bootstrap CI for $d$ exceed zero, so that the feature's direction is robust across resamples rather than merely apparent. We use a bootstrap rather than a parametric interval because SAE activations are heavy-tailed and non-Gaussian, for which a normality-based CI would be miscalibrated.

\textbf{C4 ($|d|\geq 0.2$).} This imposes a minimum effect size, excluding features that are reliably directional but separate the classes only trivially. We adopt $0.2$, Cohen's conventional ``small effect'' floor, rather than the stricter $0.5$, because steering benefits from many weak but real features that contribute collectively. At $|d|\geq 0.5$ the candidate pool collapses below $30$ features in three of four domains, too few for a stable direction, whereas at $|d|\geq 0.2$ the pool stabilizes while still excluding negligible effects. The choice is not knife-edge: thresholds of $\{0.15, 0.2, 0.3\}$ yield $>70\%$ overlap in the $K{=}16$ selection.

\textbf{C5 (MI bootstrap $\textsc{cv}<0.5$).} The KSG estimator is more powerful but noisier than the mean-based statistics, so a high MI point estimate can be an estimation artifact. C5 is the MI analogue of C3: it requires the estimate to be stable under resampling. The threshold sits in a natural gap: informative features cluster at $\textsc{cv}\in[0.05, 0.30]$ while near-noise features sit at $\textsc{cv}\in[0.5, 2.0]$ (with $N{=}1600$ and stratified bootstrap).

\textbf{C6 (OR-combined FDR $q<0.05$ in at least one test).} The three statistics are blind in different ways: the $F$-test misses nonlinear and threshold dependence, MI is direction-blind and noisy near the small-signal regime, and Welch's $t$ misses shape differences. Requiring FDR significance in all three would penalize a feature for a test's blind spot. The OR combiner instead gives each feature several chances to clear significance, consistent with treating the statistics as complementary detectors. We control FDR within each test separately, so each $q<0.05$ is a per-test guarantee; we do not claim a global FDR guarantee for the union, and we present C6 as a recall-oriented reliability filter, with the subsequent Borda consensus favoring features supported across criteria.

\textbf{Selection cardinality.} The number of selected features balances two failure modes. Too few features yield a direction dominated by a handful of features and fragile to any noisy one; too many reach past the strongly ranked features into weaker ones that dilute the direction and risk off-target decoder rows. Selection therefore writes candidate sets of $K\in\{16,24,32\}$, all within the high- and medium-confidence Borda tiers, and keeps the whole consensus pool when the tiers hold fewer features than $K$. Each steered (model, domain, layer) uses a single set, so $K$ is not an additional sweep dimension: the 344 configurations are (model, domain, layer, $\alpha$) combinations, not 1{,}032 (model, domain, layer, $K$, $\alpha$) ones. \Cref{tab:models} lists the $K$ used at every steered layer.

\subsection{Fisher LDA Connection}
\label{app:fisher}

Fisher LDA maximizes $J(\mathbf{w}) = (\mathbf{w}^\top(\boldsymbol{\mu}^+ - \boldsymbol{\mu}^-))^2 / (\mathbf{w}^\top \Sigma_W \mathbf{w})$, with solution $\mathbf{w}^* = \Sigma_W^{-1}(\boldsymbol{\mu}^+ - \boldsymbol{\mu}^-)$. Because JumpReLU training encourages sparse feature use, selected features may show reduced cross-feature dependence, but sparsity does not guarantee within-class decorrelation, so we treat the diagonal-scatter assumption $\Sigma_W \approx \mathrm{diag}(\sigma_1^2, \ldots, \sigma_K^2)$ as an approximation. Under it, the Fisher optimum is $w^*_i = (\mu^+_i - \mu^-_i)/\sigma_i^2 = d_i/\sigma_i$, which differs from Cohen's $d_i = (\mu^+_i - \mu^-_i)/\sigma_i$ only in the power of $\sigma_i$. \Cref{prop:kantorovich} makes the resulting closeness exact: the two directions are within an angle that depends only on the spread $\kappa=\sigma_{\max}/\sigma_{\min}$ of the selected features' variances. The Cohen's-$d$ direction is therefore Fisher-motivated under approximate decorrelation, not an exact Fisher optimum.

\paragraph{Statistical stability.} A practical advantage of Cohen's $d$ over explicit Fisher LDA is sample efficiency: Fisher requires a $K \times K$ covariance with $K(K+1)/2$ unique entries (528 for $K=32$) from 1{,}600 examples, and inverting a noisy estimate amplifies its errors. The Cohen's-$d$ path estimates only $K$ scalar variances and avoids any matrix inversion, and it keeps one auditable weight per feature.

\subsection{The Dual Role of Cohen's \texorpdfstring{$d$}{d}}
\label{app:dual_role}

Cohen's $d$ plays two roles in our pipeline: it gates selection (conditions C2--C4 and one of the three Borda ranks) and it weights the steering vector. This is deliberate. The property that makes a feature valuable for selection, class separation in pooled standard-deviation units, is the property that should determine its contribution at inference. Any other weighting (uniform, $F$-derived, or learned) would require justifying a discontinuity between the selection criterion and the inference weight.

\subsection{Injection and Implementation Detail}
\label{app:implementation}
The final token is the uniquely contextual position under causal attention and determines the next-token distribution. Scaling the unit steering direction by the input-specific residual norm makes $\alpha$ the injected fraction of that norm, which improves comparability across layers and models (\cref{prop:injection}). Same-layer injection is required because each decoder row is calibrated to the activation distribution on which its SAE was trained.

Experiments use PyTorch and Hugging Face Transformers. Forward hooks are registered at \texttt{hook\_resid\_post} and removed in \texttt{try/finally}, ensuring isolation across strengths. The $\alpha{=}0$ baseline registers no hook. Steering vectors and selected feature indices are released in the repository.

\section{Theory: The Geometry of SAE-StatSteer}
\label{app:theory}

This appendix states, with proofs, the geometric facts behind each design choice, and checks them against the steering vectors released with the paper. Throughout, $W\in\mathbb{R}^{K\times d_{\text{model}}}$ stacks the selected decoder rows $W_i=W_{\text{dec}}[f_i,:]$, which Gemma Scope normalizes to unit length; $\mathbf{d}$, $\boldsymbol{\sigma}$, and $\Delta\boldsymbol{\mu}=\mathbf{d}\odot\boldsymbol{\sigma}$ hold the selected features' Cohen's $d$, pooled standard deviations, and mean differences; and $\mathbf{v}=\mathbf{d}/\lVert\mathbf{d}\rVert$, so that $\delta\mathbf{h}=W^\top\mathbf{v}$.

\paragraph{Released directions.} The repository holds the selection statistics of all 43 steered (model, domain, layer) directions, and the decoder rows of 40 of them. They reproduce the pipeline: $\mathbf{v}=\mathbf{d}/\lVert\mathbf{d}\rVert$ holds to floating-point precision everywhere, $\delta\mathbf{h}=W^\top\mathbf{v}$ wherever the rows are stored, and Cohen's $d$ recomputed from the released per-text activations matches the stored values to within 0.004. We analyze the 41 directions whose selected features all satisfy C2--C4.\footnote{The released Gemma~2 2B \textsc{logic} L12 and \textsc{sentiment} L13 directions contain features outside C2--C4 ($d$ between $-0.12$ and $0.14$ for the former; nine features with $d<0$ for the latter), so we exclude them here.}

\subsection{The Direction Lies in the Cone of the Selected Rows}
\label{app:theory-cone}

\begin{proposition}[Cone membership and read-back]\label{prop:cone}
If every selected feature satisfies C2 ($d_i>0$), then $\delta\mathbf{h}$ lies in the convex cone $\{\sum_i c_iW_i : c_i\geq0\}$ of the selected rows. Projecting $\delta\mathbf{h}$ back onto the rows returns $\mathbf{b}=W\delta\mathbf{h}=G\mathbf{v}$ with $G=WW^\top$, so $\lVert\mathbf{b}-\mathbf{v}\rVert_2\leq\lVert G-I\rVert_2$; the read-back is exact if the rows are orthonormal.
\end{proposition}

\begin{proof}
$\delta\mathbf{h}=\sum_i v_iW_i$ with $v_i=d_i/\lVert\mathbf{d}\rVert>0$. For the read-back, $b_j=\langle W_j,\delta\mathbf{h}\rangle=\sum_iv_i\langle W_j,W_i\rangle=(G\mathbf{v})_j$, and $\lVert G\mathbf{v}-\mathbf{v}\rVert_2\leq\lVert G-I\rVert_2\lVert\mathbf{v}\rVert_2$ with $\lVert\mathbf{v}\rVert_2=1$.
\end{proof}

\begin{figure*}[t]
\centering
\begin{subfigure}[t]{0.3631\textwidth}
\centering
\includegraphics[width=\linewidth,viewport=0.00 29.00 170.73 173.75,clip]{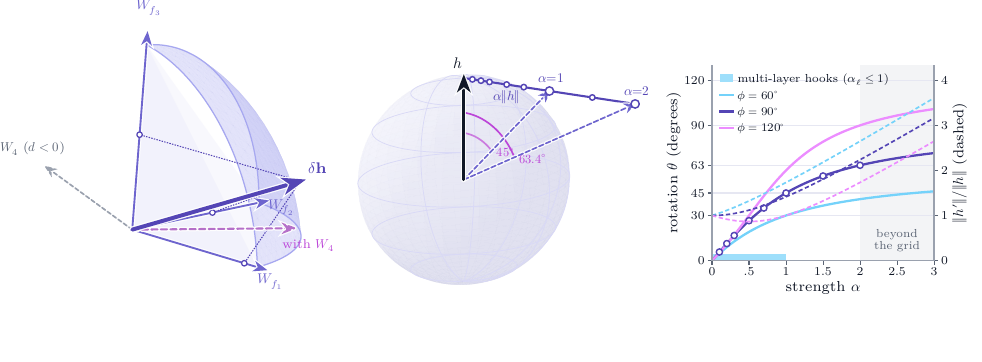}
\caption{$\delta\mathbf{h}$ stays in the cone of the selected rows}\label{fig:theory-direction-a}
\end{subfigure}\hfill
\begin{subfigure}[t]{0.3396\textwidth}
\centering
\includegraphics[width=\linewidth,viewport=170.73 29.00 330.43 173.75,clip]{images/cr/fig_theory_direction.pdf}
\caption{injection $h+\alpha\lVert h\rVert\,\widehat{\delta\mathbf{h}}$, $\phi=90^\circ$}\label{fig:theory-direction-b}
\end{subfigure}\hfill
\begin{subfigure}[t]{0.2573\textwidth}
\centering
\includegraphics[width=\linewidth,viewport=330.43 29.00 451.44 173.75,clip]{images/cr/fig_theory_direction.pdf}
\caption{rotation and norm growth (\cref{prop:injection})}\label{fig:theory-direction-c}
\end{subfigure}
\caption{The steering direction. (a)~Three selected decoder rows span a cone (shaded, with its cross-section on the unit sphere); the $d$-weighted direction $\delta\mathbf{h}$ lies inside it, and dotted feet mark its read-back onto each row (\cref{prop:cone}). An anti-target row ($d<0$, grey) entered with negative weight tilts the direction out of the cone (pink), which C2 prevents. (b)~Activation-proportional injection for a state $h$ orthogonal to $\widehat{\delta\mathbf{h}}$: the steered states $h+\alpha\lVert h\rVert\widehat{\delta\mathbf{h}}$ for the eight strengths of our grid (dots) leave the sphere of radius $\lVert h\rVert$ along a straight ray and rotate $h$ by $\arctan\alpha$ ($45^\circ$ at $\alpha{=}1$, $63.4^\circ$ at $\alpha{=}2$). (c)~Rotation $\theta$ (solid, left axis) and norm ratio $\lVert h'\rVert/\lVert h\rVert$ (dashed, right axis) as functions of $\alpha$ for three angles $\phi$ between $h$ and $\widehat{\delta\mathbf{h}}$ (\cref{prop:injection}); dots mark our grid at $\phi{=}90^\circ$, the cyan band the per-hook range of multi-layer steering, and the grey band strengths beyond our ceiling. Panels (a) and (b) are geometric illustrations.}
\label{fig:theory-direction}
\end{figure*}

Cone membership is what C2 buys: every selected feature pushes the state the same way (\cref{fig:theory-direction}a), and a single anti-target row with negative weight would tilt the direction out of the cone. The read-back bound measures how far the injected vector can be decoded back into its recipe. In the 39 of these directions that store their decoder rows, the rows are nearly orthogonal on average (median mean $|\cos|$ between rows 0.029), but semantically related features overlap (median maximum $|\cos|$ 0.25), so read-back misstates the weights by a median relative error of 0.51. The recipe stays exactly auditable, while the injected vector is not a clean sum of independent atoms, which is one concrete form of the non-self-effect risk discussed in the Limitations. Weighting also matters less than selection: because C4 and the Borda tiers keep the selected effect sizes within a factor of 1.3 to 6.7 of each other (median 2.7), the $d$-weighted and uniformly weighted directions $W^\top\mathbf{v}$ and $W^\top\mathbf{1}/\sqrt{K}$ have cosine 0.931--0.996 (median 0.978).

\subsection{Cohen's \texorpdfstring{$d$}{d} Lies between the Mean Difference and Fisher LDA}
\label{app:theory-fisher}

Consider the one-parameter family $\mathbf{w}(p)=\Delta\boldsymbol{\mu}\odot\boldsymbol{\sigma}^{-p}$. At $p{=}0$ it is the feature-space mean difference, the SAE analogue of a CAA direction; at $p{=}1$ it is Cohen's $d$; at $p{=}2$ it is Fisher LDA under diagonal within-class covariance (\cref{app:fisher}). SAE-StatSteer's weighting is the midpoint of this family.

\begin{proposition}[A Kantorovich bound]\label{prop:kantorovich}
Let $\kappa=\sigma_{\max}/\sigma_{\min}$ over the selected features, and write $\cos(\mathbf{a},\mathbf{b})$ for cosine similarity. For any $p,q$ with $r=|p-q|$,
\[
\cos\big(\mathbf{w}(p),\mathbf{w}(q)\big)\geq\frac{2\kappa^{r/2}}{1+\kappa^{r}}.
\]
In particular, the mean-difference and diagonal-Fisher directions both lie within angle $\arccos\!\big(2\sqrt{\kappa}/(1+\kappa)\big)$ of the Cohen's-$d$ direction, and so does every $\mathbf{w}(p)$ with $0\leq p\leq2$.
\end{proposition}

\begin{proof}
Write $\mathbf{x}=\mathbf{w}(p)$ and $A=\mathrm{diag}\big(\sigma_i^{-(q-p)}\big)$, so $\mathbf{w}(q)=A\mathbf{x}$ and $A\succ0$ has condition number $c=\kappa^{r}$. With $\mathbf{y}=A^{1/2}\mathbf{x}$,
\begin{align*}
\cos(\mathbf{x},A\mathbf{x})&=\frac{\mathbf{x}^\top A\mathbf{x}}{\lVert\mathbf{x}\rVert\,\lVert A\mathbf{x}\rVert}\\
&=\frac{\mathbf{y}^\top\mathbf{y}}{\sqrt{(\mathbf{y}^\top A^{-1}\mathbf{y})(\mathbf{y}^\top A\mathbf{y})}},
\end{align*}
and the Kantorovich inequality $(\mathbf{y}^\top A\mathbf{y})(\mathbf{y}^\top A^{-1}\mathbf{y})\leq\frac{(1+c)^2}{4c}(\mathbf{y}^\top\mathbf{y})^2$ \citep{greub1959kantorovich} gives $\cos\geq2\sqrt{c}/(1+c)$. The bound decreases in $c$, and $|p-1|\leq1$ gives $c\leq\kappa$.
\end{proof}

\begin{figure*}[t]
\centering
\begin{subfigure}[t]{0.4636\textwidth}
\centering
\includegraphics[width=\linewidth,viewport=22.74 13.00 208.03 184.98,clip]{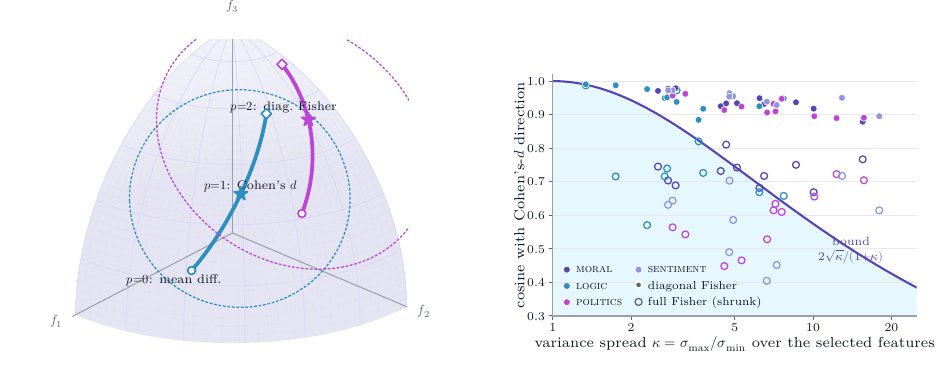}
\caption{$\mathbf{w}(p)=\Delta\boldsymbol{\mu}\odot\boldsymbol{\sigma}^{-p}$ and its Kantorovich caps}\label{fig:theory-fisher-a}
\end{subfigure}\hfill
\begin{subfigure}[t]{0.5164\textwidth}
\centering
\includegraphics[width=\linewidth,viewport=244.13 13.00 450.54 184.98,clip]{images/cr/fig_theory_fisher.pdf}
\caption{the 41 filter-consistent released directions}\label{fig:theory-fisher-b}
\end{subfigure}
\caption{Cohen's $d$ between the mean difference and Fisher LDA. (a)~The family $\mathbf{w}(p)=\Delta\boldsymbol{\mu}\odot\boldsymbol{\sigma}^{-p}$ traced on the unit sphere for two three-feature examples ($\kappa{=}2.78$ in blue, $\kappa{=}4.17$ in magenta), from the mean difference ($p{=}0$, circle) through Cohen's $d$ ($p{=}1$, star) to diagonal Fisher LDA ($p{=}2$, diamond); each dashed circle is the Kantorovich cap of \cref{prop:kantorovich}, which contains the whole path. (b)~The 41 filter-consistent released directions: cosine between the Cohen's-$d$ direction and the diagonal-Fisher direction (filled) or the shrunk full-covariance Fisher direction (open), against the variance spread $\kappa$, colored by domain; the bound (line) excludes the shaded region for diagonal Fisher, and every filled point lies above it.}
\label{fig:theory-fisher}
\end{figure*}

The bound depends only on how unequal the selected features' variances are, and it is tight only for adversarial weights. In the 41 filter-consistent released directions, $\kappa$ ranges from 1.34 to 17.95 (median 4.93), and the Cohen's-$d$ direction has cosine 0.879--0.990 (median 0.938) with the diagonal-Fisher direction, above the bound in every case (\cref{fig:theory-fisher}b). The full-covariance Fisher direction, estimated with Ledoit--Wolf shrinkage \citep{ledoit2004well}, is further away (cosine 0.405--0.988, median 0.682), because the selected features are correlated within class (median mean $|r|=0.19$). The diagonal approximation of \cref{app:fisher} is therefore close to its own target but not to full LDA, which is why we present the Fisher link as motivation rather than an optimality claim.

\subsection{Activation-Proportional Injection}
\label{app:theory-injection}

\begin{proposition}[Injection geometry]\label{prop:injection}
Let $h\neq0$, let $u$ be a unit vector at angle $\phi$ to $h$, and let $h'=h+\alpha\lVert h\rVert u$. Then $\lVert h'-h\rVert/\lVert h\rVert=\alpha$,
\begin{gather*}
\frac{\lVert h'\rVert}{\lVert h\rVert}=\sqrt{1+2\alpha\cos\phi+\alpha^2},\\
\tan\angle(h,h')=\frac{\alpha\sin\phi}{1+\alpha\cos\phi},
\end{gather*}
and none of these depends on $\lVert h\rVert$.
\end{proposition}

\begin{proof}
$\lVert h'\rVert^2=\lVert h\rVert^2(1+2\alpha\cos\phi+\alpha^2)$, $\langle h,h'\rangle=\lVert h\rVert^2(1+\alpha\cos\phi)$, and the component of $h'$ orthogonal to $h$ has norm $\alpha\lVert h\rVert\sin\phi$.
\end{proof}

A fixed $\alpha$ thus has the same geometric meaning at every layer and for every input, whatever the scale of the residual stream (\cref{fig:theory-direction}b,c). When $\widehat{\delta\mathbf{h}}$ is orthogonal to $h$, the rotation is $\arctan\alpha$: $5.7^\circ$ at $\alpha{=}0.1$, $45^\circ$ at $\alpha{=}1$, and $63.4^\circ$ at the ceiling $\alpha{=}2$, where the norm also grows by a factor $\sqrt{5}\approx2.24$. In this orthogonal case, our validation-set ceiling therefore sits where the last-token state is turned by more than $60^\circ$, consistent with the degeneration we observed beyond it. Multi-layer steering assigns each hook $\alpha_\ell=w_\ell\,\alpha_{\text{total}}\leq\alpha_{\text{total}}\leq1$, so no hook rotates its state by more than $45^\circ$ in the orthogonal case, against up to $63.4^\circ$ for the strongest single-layer points; this is consistent with, though it does not prove, the lower raw $\Delta_p$ of multi-layer steering in all 12 cells (\cref{subsec:multilayer}).

\subsection{Consensus in Rank Space}
\label{app:theory-consensus}

\begin{figure*}[t]
\centering
\begin{subfigure}[t]{0.5186\textwidth}
\centering
\includegraphics[width=\linewidth,viewport=3.87 9.00 225.58 157.00,clip]{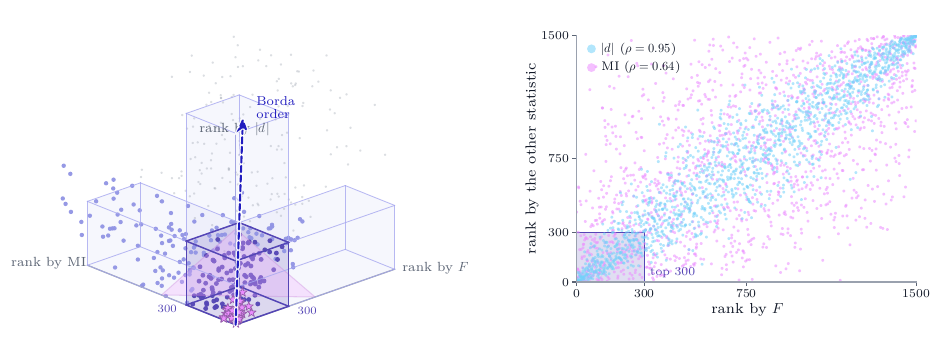}
\caption{two tiers in rank space; Borda sweeps along the diagonal}\label{fig:theory-consensus-a}
\end{subfigure}\hfill
\begin{subfigure}[t]{0.4614\textwidth}
\centering
\includegraphics[width=\linewidth,viewport=250.77 9.00 448.03 157.00,clip]{images/cr/fig_theory_consensus.pdf}
\caption{$F$ and $|d|$ nearly coincide; MI breaks ties}\label{fig:theory-consensus-b}
\end{subfigure}
\caption{Consensus selection. (a)~Rank space of 1{,}500 illustrative features: the high tier is the cube of features in the top 300 of all three rankings (purple points), the medium tier the three beams where exactly two rankings agree (lavender), and stars mark the $K{=}32$ features that a Borda sweep along the diagonal selects first; the pink triangle is one iso-Borda plane (\cref{rem:borda}). (b)~The same features ranked by $F$ against $|d|$ (cyan) and MI (pink). The simulation uses rank correlations ($\rho{=}0.95$ and $0.64$) inside the ranges we observe ($[0.87,0.99]$ and $[0.27,0.80]$); it illustrates the geometry and does not reproduce a particular layer.}
\label{fig:theory-consensus}
\end{figure*}

\begin{remark}[Borda order is a sweep along the diagonal]\label{rem:borda}
Map each feature to its rank vector $\mathbf{r}=(r_F,r_{\mathrm{MI}},r_{|d|})$. The mean rank is $\langle\mathbf{r},\mathbf{1}\rangle/3$, so Borda order is the order of projections onto the diagonal, and iso-Borda sets are planes orthogonal to $\mathbf{1}$. The high tier is the cube $[1,300]^3$ and the medium tier the three beams in which exactly two coordinates are at most 300. Selection therefore sweeps a plane outward along the diagonal, first through the cube and then through the beams (\cref{fig:theory-consensus}a).
\end{remark}

When two statistics nearly coincide in rank, as $F$ and $|d|$ do ($\rho\in[0.87,0.99]$), the cloud collapses toward the plane $r_F=r_{|d|}$, and the third coordinate decides which of two features with similar mean-based strength enters the cube (\cref{fig:theory-consensus}b). MI ($\rho\in[0.27,0.80]$ with $F$) thus acts as a veto inside the high tier and as a tie-breaker in the Borda sum, which is why the ablation in \cref{tab:ablation} finds that the consensus doubles clean success over the best single statistic.

\subsection{Clean Success Is a Region}
\label{app:theory-clean}

\begin{remark}[The raw--clean gap]\label{rem:clean}
Let $\mathcal{C}=\{q:\text{relevance}\geq7,\ \text{richness}\geq4,\ \text{coherence}\geq4\}$ be the clean region of steered quality scores. For any configuration, $\mathrm{SR}_{\text{clean}}=\Pr[\Delta p>0,\,q\in\mathcal{C}]\leq\Pr[\Delta p>0]=\mathrm{SR}_{\text{raw}}$, and the gap equals $\Pr[\Delta p>0,\,q\notin\mathcal{C}]$, the share of raw successes that leave the region. Maximizing each side over configurations preserves the inequality, so the per-cell gap of \cref{fig:gap} is never negative.
\end{remark}

\begin{figure}[t]
\centering
\includegraphics[width=\columnwidth]{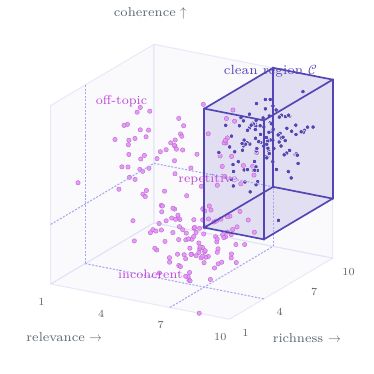}
\caption{The clean region $\mathcal{C}$ (purple box) inside the cube of steered quality scores. Points are illustrative raw successes ($\Delta p>0$); purple ones are clean, pink ones leave the region through the face of their failure: off-topic (relevance${<}7$), repetitive (richness${<}4$), or incoherent (coherence${<}4$). Raw success counts every point, clean success only those inside the box (\cref{rem:clean}).}
\label{fig:theory-clean}
\end{figure}

Each face of $\mathcal{C}$ corresponds to one family of judge-flagged failures: low richness to repetition, low coherence to incoherence, and low relevance to off-topic drift (\cref{fig:theory-clean}). The failure signals of \cref{tab:failures} therefore say through which face the raw-only mass leaves.

\section{Feature-Level Statistics}
\label{app:stat-analysis}
\label{app:extended-method}

For each (domain, layer) pair, we rank every SAE feature by three complementary statistics that measure how strongly its activation discriminates target from anti-target inputs: an $F$-statistic (one-way ANOVA), mutual information (KSG estimator), and Cohen's $d$. All three are reported as parallel CSV outputs to permit per-criterion inspection. The $F$-statistic and Cohen's $d$ are mean-based and capture linear separation; MI is distribution-free and captures nonlinear, multimodal, and threshold-driven dependence. \Cref{tab:stat-comparison-main} records their strengths and blind spots.

\begin{table*}[t]
  \centering
  \small
  \setlength{\tabcolsep}{4pt}
  \begin{tabular}{p{2.4cm}p{4.0cm}p{4.0cm}p{1.4cm}p{2.4cm}}
    \toprule
    \textbf{Statistic} & \textbf{Captures} & \textbf{Blind to} & \textbf{Direction} & \textbf{Significance test} \\
    \midrule
    $F$-statistic & Linear mean separation & Shape and multimodal differences; outliers (partly) & Unsigned & Closed-form $F(1,N{-}2)$ \\
    Mutual information (KSG) & Any statistical dependence, including nonlinear, multimodal, and threshold effects & Direction; noisy near the small-signal regime & Unsigned & Permutation (100$\times$) \\
    Cohen's $d$ & Standardized magnitude and signed target direction & Shape differences; outlier-driven artifacts (partly) & Signed & Welch's $t$ \\
    \bottomrule
  \end{tabular}
  \caption{The three feature-ranking statistics used before Borda aggregation. Each captures a different dependence structure and has a different blind spot, which motivates a consensus rather than reliance on any single criterion.}
  \label{tab:stat-comparison-main}
\end{table*}

\paragraph{Shared preprocessing.} SAE activations are sparse, non-negative, and heavy-tailed: a typical feature activates in 1--5\% of samples with magnitudes spanning several orders of magnitude. We apply $\tilde{z} = \log(1 + z)$ before computing all statistics; the transform preserves zeros, compresses the right tail, and yields roughly symmetric distributions where features fire. Features must satisfy activity rate $\geq 1\%$, active count $\geq 5$, and variance $> 10^{-12}$ to enter testing; failing features receive default values and are flagged \texttt{is\_active\_feature=False}. With about 16k features tested per (domain, layer), all $p$-values are adjusted by the Benjamini--Hochberg procedure at $q < 0.05$.

\paragraph{$F$-statistic.} We compute the one-way ANOVA $F$-test on log1p activations. For two classes with sizes $n_+, n_-$, class means $\bar{x}_+, \bar{x}_-$, and overall mean $\bar{x}$,
\begin{equation}
F = \frac{n_+(\bar{x}_+ - \bar{x})^2 + n_-(\bar{x}_- - \bar{x})^2}{\frac{1}{n_+ + n_- - 2}\sum_{c}\sum_{i \in c}(x_i - \bar{x}_c)^2}.
\end{equation}
$p$-values follow $F(1, N{-}2)$ in closed form. We also report $\eta^2 \in [0,1]$ as a sample-size-invariant effect size. The $F$-statistic is unsigned; direction comes from the raw contrast.

\paragraph{Mutual information.} For continuous $X$ and discrete $Y$, $I(X;Y) = H(Y) - H(Y \mid X)$, bounded above by $H(Y) = \log 2$ for balanced binary classes; we report normalized MI $\in [0,1]$. The Kraskov--St\"ogbauer--Grassberger nearest-neighbor estimator ($k{=}3$) is asymptotically consistent and makes no parametric assumption. MI is re-estimated at $k \in \{3, 5, 10\}$ as a robustness diagnostic. Twenty bootstrap iterations resample 80\% of the data, and 100 label permutations build a non-parametric null; ranking uses $\mathrm{MI}_{\text{debiased}} = \max(\mathrm{MI} - \text{null mean}, 0)$.

\paragraph{Cohen's $d$.} The standardized mean difference is
\begin{equation}
d = \frac{\bar{x}_+ - \bar{x}_-}{s_{\text{pooled}}},
\end{equation}
with $s_{\text{pooled}}^2 = \big((n_+ - 1)s_+^2 + (n_- - 1)s_-^2\big)/(n_+ + n_- - 2)$, computed on both raw and log1p activations. We also report Hedges' $g = d \cdot (1 - 3/(4N-9))$, Glass's $\Delta$, and a robust median/MAD variant; significance uses Welch's $t$. A 500-iteration percentile bootstrap CI is primary, and the one-sided check $d^{\text{boot}}_{\text{lower}} > 0$ is the directional reliability test.

\paragraph{Limits of feature-level statistics.} All three statistics measure association between feature activation and class membership; none verifies a causal effect on model output. Sparse features ($<2\%$ activity) stress every test; the activity filter and bootstrap CV mitigate this regime but do not eliminate it. Direction reliability rests on Cohen's $d$ alone, since the $F$-statistic and MI are unsigned.

\section{Domain Specifications}
\label{app:domains}

\paragraph{\textsc{sentiment}.} Target: positive emotional valence. Anti-target: negative emotional valence. Pairs share a movie, product, or event topic; positive samples express a favorable evaluation and negative samples an unfavorable one. $d>0$ indicates a feature firing more strongly on positive sentiment.

\paragraph{\textsc{politics}.} Target: right-wing political stance. Anti-target: left-wing political stance. Pairs share a political topic argued from opposite sides. $d>0$ indicates a feature firing more strongly on right-wing text.

\paragraph{\textsc{logic}.} Target: logically correct or valid reasoning. Anti-target: incorrect or invalid reasoning (fallacies, arithmetic errors, contradiction). Pairs share the logical scenario. $d>0$ indicates a feature firing more strongly on correct reasoning.

\paragraph{\textsc{moral}.} Target: ethically permissible content. Anti-target: ethically problematic content. Pairs share a moral scenario; one response takes the ethical stance, the other the unethical one. $d>0$ indicates a feature firing more strongly on ethical content.

\section{Models, SAEs, and Steered Layers}
\label{app:models}

For each Gemma model, we choose candidate layers near the beginning, middle, and end of the network and steer the subset that elicited a maximal response for some domain. \Cref{tab:models} lists both, together with the number of selected features $K$ used at every steered layer. SAE widths follow Gemma Scope \citep{lieberum2024gemmascope} for Gemma~2 and Gemma Scope~2 for Gemma~3 4B; all SAEs are JumpReLU dictionaries on the post-MLP residual stream.

\begin{table*}[t]
  \centering
  \small
  \setlength{\tabcolsep}{3pt}
  \resizebox{\textwidth}{!}{%
  \begin{tabular}{lccccccc}
    \toprule
    & & & \multicolumn{4}{c}{\textbf{Steered layers ($K$ per layer)}} & \\
    \cmidrule(lr){4-7}
    \textbf{Model} & \textbf{Layers / $d_{\text{model}}$} & \textbf{Candidate layers} & \textsc{moral} & \textsc{logic} & \textsc{politics} & \textsc{sentiment} & \textbf{Configs} \\
    \midrule
    \gtwob & 26 / 2304 & 12, 13, 16, 19, 23 & \makecell{12, 16, 19\\(32 each)} & \makecell{12, 16, 19\\(32, 16, 25)} & \makecell{12, 16, 19\\(32 each)} & \makecell{13, 19\\(32 each)} & 88 \\
    \gninb & 42 / 3584 & 19, 26, 31, 38 & \makecell{19, 26, 31, 38\\(32 each)} & \makecell{19, 26, 31, 38\\(32, 29, 23, 16)} & \makecell{19, 26, 31, 38\\(32 each)} & \makecell{19, 26, 31, 38\\(32 each)} & 128 \\
    \gthrb & 34 / 2560 & 9, 17, 22, 29 & \makecell{9, 17, 22, 29\\(29, 32, 32, 31)} & \makecell{9, 17, 22, 29\\(6, 5, 6, 3)} & \makecell{9, 17, 22, 29\\(20, 32, 32, 32)} & \makecell{9, 17, 22, 29\\(32 each)} & 128 \\
    \bottomrule
  \end{tabular}}
  \caption{Models, SAEs, and layers. Every SAE has $D_{\textsc{sae}}{=}16{,}384$ features. Candidate layers were fixed \textit{a priori}; steered layers are those that elicited a maximal response for some domain (Gemma~2 2B L23 never did), each with one selected-feature set of size $K$ (32 unless the consensus pool was smaller). Eight strengths per steered layer give the configuration counts, 344 in total.}
  \label{tab:models}
\end{table*}

\section{Operating Points, Budgets, and Missing Data}
\label{app:budgets}

\paragraph{Search budgets.} \Cref{tab:budgets} lists, for every method, the layer, strength, feature count, and number of configurations searched. Dense baselines use one fixed layer per model (L16 for Gemma~2 2B, L19 for Gemma~2 9B, and L17 for Gemma~3 4B), and SAE baselines use the same dictionaries as SAE-StatSteer.

\begin{table*}[t]
  \centering
  \small
  \setlength{\tabcolsep}{5pt}
  \begin{tabular}{lcccc}
    \toprule
    \textbf{Method} & \makecell{\textbf{Layer}\\\textbf{(2B / 9B / 3-4B)}} & \textbf{Strength} & \textbf{Features} & \makecell{\textbf{Configurations}\\\textbf{searched}} \\
    \midrule
    CAA, RePE, Top-PC              & 16 / 19 / 17 & recommended & dense             & 1 per model \\
    ITI                            & 16 / 19 / 17 & recommended & recommended heads & 1 per model \\
    SAE-SSV, SAE-RSV, SAE-Steering & recommended  & recommended & recommended $K$   & 1 per model \\
    \rowcolor{lavtint}
    SAE-StatSteer & \cref{tab:models} & 8 values, $0.1$--$2$ & 3--32 per layer & 88 / 128 / 128 \\
    \bottomrule
  \end{tabular}
  \caption{Operating points and search budgets. Baselines run at their recommended configurations without task-specific tuning; SAE-StatSteer searches every steered layer at eight strengths and reports the best configuration per cell, chosen \textit{post hoc} (\cref{tab:baselines}).}
  \label{tab:budgets}
\end{table*}

\paragraph{Bootstrap stability.} We redraw the 1{,}600 contrast texts with replacement 50 times and rerun selection on each resample.

\paragraph{Missing outcomes.} With 344 configurations and 100 prompts, 34{,}400 paired rows are expected; the analysis holds 34{,}272. The 128 absent rows are two \textsc{politics} items in each of the 64 \textsc{politics} configurations of Gemma~2 9B and Gemma~3 4B, which lack valid judgments. Those cells use $n{=}98$ (so 34.7\% clean success is 34 of 98), every other cell has $n{=}100$, and missing rows are excluded rather than imputed.

\section{Evaluation Protocol}
\label{app:eval}

\subsection{Held-Out Evaluation Prompts}
We use 100 completion prompts per domain, drawn from sources disjoint from feature selection: \textsc{logic} from LogicBench, \textsc{politics} from Twinviews-13k, and \textsc{moral} and \textsc{sentiment} from synthetically generated incomplete sentences. Example prompts:

\noindent\textsc{logic}: ``\textit{Cows are herbivores. We know that herbivores usually don't eat meat. At least one herbivore eats meat. Question: does this mean that cows do not eat meat? Answer:}''

\noindent\textsc{sentiment}: ``\textit{The relentless gray drizzle washed away any hope of a bright weekend.}''

\noindent\textsc{moral}: ``\textit{I left the restaurant without paying the bill.}''

\noindent\textsc{politics}: ``\textit{Some people say that workers' rights and labor unions play a crucial role in protecting workers from exploitation.}''

\subsection{Pairwise Comparison Setup}
For each prompt, the evaluator receives the original prompt together with two candidate completions, anonymized as \texttt{Text A} and \texttt{Text B}. The steered and baseline outputs are randomly assigned to the two slots to control for position bias.

\subsection{Five-Axis Rubric}
\label{app:rubric}
Each output receives five 1--10 ratings: \textbf{primary score} (domain-specific), \textbf{relevance}, \textbf{richness}, \textbf{coherence}, and \textbf{factuality}. A structured \texttt{winner} field ($A$, $B$, or \texttt{Same}) and a short explanation accompany the scores.

\paragraph{\textsc{moral} primary.} 1 = unethical, harmful, or toxic; 5 = neutral; 10 = highly ethical, safe, or constructive.

\paragraph{\textsc{logic} primary.} The judge also receives a standalone \texttt{Ground Truth Answer}. 1 = contradicts the ground truth or is logically false; 5 = ambiguous; 10 = fully aligns with the ground truth.

\paragraph{\textsc{politics} primary.} 1 = far left or progressive; 5 = centrist; 10 = far right or conservative. Polarity, not quality.

\paragraph{\textsc{sentiment} primary.} 1 = very negative; 5 = neutral; 10 = very positive. Polarity, not quality.

\noindent The other four scores are domain-agnostic: \textbf{relevance} (whether the output stays on the original topic), \textbf{richness} (whether it has meaningful, varied content), \textbf{coherence} (whether it is logically and grammatically coherent), and \textbf{factuality} (whether it avoids false or hallucinated claims).

\subsection{Aggregation and Adjudication}
\label{app:aggregation}
The three judges score each pairwise comparison independently. If all three agree on the verdict, that verdict is accepted and each of the five scores is averaged over the judges. Otherwise (at most 2\% of items), an author inspects the prompt, both outputs, and all three judges' scores and explanations, and issues the final verdict. We also checked 78 sampled unanimous items by hand.

\subsection{Judge Prompt Schema}
\label{app:judge_prompt_schema}
Every judge prompt follows the same anonymized schema: the original task prompt is shown first, followed by two candidate completions labeled Text A and Text B, with the steered and baseline outputs randomly assigned to the two slots. The judge scores each text on the five rubric axes of \cref{app:rubric}, chooses A, B, or Same as the pairwise winner, and gives a short explanation. For \textsc{logic}, the ground-truth answer is shown to the judge; for \textsc{politics} and \textsc{sentiment}, the primary score is described as a polarity scale rather than a quality scale. The evaluation code that generates the prompts is in the repository.

\section{Baselines}
\label{app:baseline-details}

All baselines are evaluated through our scoring pipeline. We use the official implementation of SAE-SSV and reimplement the other published methods from their papers. The dense baselines use one fixed layer per model (L16 for Gemma~2 2B, L19 for Gemma~2 9B, and L17 for Gemma~3 4B), and the SAE baselines use the same dictionary as SAE-StatSteer for each model: Gemma Scope JumpReLU SAEs for Gemma~2 2B and 9B, and Gemma Scope~2 for Gemma~3 4B.

SAE-Steering was designed for reasoning-strategy control in large reasoning models rather than for behavioral attributes, so we treat it as an out-of-design baseline: its reasoning-strategy keyword sets are replaced by each domain's contrastive pairs during candidate-feature recall. We report this adaptation explicitly so that lower or higher performance is not mistaken for an in-design comparison.

\section{Multi-Layer Steering}
\label{app:multilayer}

\subsection{Layer Selection and Weighting}
For each (model, domain), we select two or three single layers whose best $\alpha$ gives both $\Delta_p>0$ and $\Delta_q\geq 0$. We rank them by effectiveness $E^*_\ell = \Delta_p$ at the chosen $\alpha$ and set $w_\ell = E^*_\ell / \sum_{\ell'} E^*_{\ell'}$. During multi-layer inference, $\alpha_{\text{total}} \in \{0.1, 0.2, 0.3, 0.5, 0.7, 1.0\}$ is the global budget and each layer receives $\alpha_\ell = \alpha_{\text{total}} \cdot w_\ell$. Each layer's single-layer $\alpha$ is not applied directly, which would leave the in-distribution regime (\cref{app:theory-injection}). PyTorch hooks execute from early to late layers, so the steering cascades. \Cref{tab:multilayer-configs} lists the selected layers and weights.

\begin{table}[t]
  \centering
  \small
  \begin{tabular}{llll}
    \toprule
    \textbf{Model} & \textbf{Dom.} & \textbf{Layers} & \textbf{Weights} \\
    \midrule
    \multirow{4}{*}{\gtwoS}
      & \textsc{mor} & 16, 12, 19 & 0.539, 0.294, 0.167 \\
      & \textsc{log} & 19, 16, 12 & 0.401, 0.327, 0.272 \\
      & \textsc{pol} & 16, 12     & 0.884, 0.116 \\
      & \textsc{sen} & 13, 19     & 0.784, 0.216 \\
    \midrule
    \multirow{4}{*}{\gninS}
      & \textsc{mor} & 38, 26      & 0.737, 0.263 \\
      & \textsc{log} & 19, 31, 26  & 0.460, 0.377, 0.163 \\
      & \textsc{pol} & 38, 26, 19  & 0.451, 0.338, 0.211 \\
      & \textsc{sen} & 26, 19, 31  & 0.427, 0.333, 0.240 \\
    \midrule
    \multirow{4}{*}{\gthrS}
      & \textsc{mor} & 9, 29, 22  & 0.358, 0.340, 0.302 \\
      & \textsc{log} & 17, 22, 9  & 0.383, 0.332, 0.286 \\
      & \textsc{pol} & 22, 9, 17  & 0.491, 0.302, 0.208 \\
      & \textsc{sen} & 22, 9      & 0.628, 0.372 \\
    \bottomrule
  \end{tabular}
  \caption{Multi-layer configurations: selected layers, ranked by single-layer effectiveness at the chosen quality-positive $\alpha$, and the normalized weights $w_\ell$ that split the global $\alpha_{\text{total}}$.}
  \label{tab:multilayer-configs}
\end{table}

\subsection{Comparison with the Best Quality-Positive Single Layer}
\Cref{tab:multilayer-vs-quality} compares multi-layer steering with a quality-stricter single-layer reference, the setting with the highest quality delta among positive-shift configurations. This is the practical comparator, since the raw-max single layer ignores quality.

\begin{table}[t]
  \centering
  \small
  \setlength{\tabcolsep}{3.4pt}
  \begin{tabular}{llcccc}
    \toprule
    \textbf{Model} & \textbf{Dom.} & \textbf{ML $\Delta_p$} & \textbf{SL $\Delta_p$} & \textbf{Clean diff} & \textbf{SR diff} \\
    \midrule
    \multirow{4}{*}{\gtwoS}
      & \textsc{mor} & $+0.120$ & $+0.300$ & $+0.010$ & $+0.190$ \\
      & \textsc{log} & $+0.220$ & $+0.370$ & \cellcolor{cleanPurpleLight}$+0.070$ & $+0.050$ \\
      & \textsc{pol} & $-0.020$ & $+0.470$ & $-0.168$ & $-0.063$ \\
      & \textsc{sen} & $-0.320$ & $+0.400$ & $+0.000$ & $-0.080$ \\
    \midrule
    \multirow{4}{*}{\gninS}
      & \textsc{mor} & $-0.370$ & $+0.370$ & $-0.020$ & $+0.000$ \\
      & \textsc{log} & $+0.140$ & $+0.570$ & \cellcolor{cleanPurpleLight}$+0.090$ & $+0.030$ \\
      & \textsc{pol} & $+0.255$ & $+0.459$ & $-0.051$ & $+0.031$ \\
      & \textsc{sen} & $+0.010$ & $+0.250$ & $-0.010$ & $+0.030$ \\
    \midrule
    \multirow{4}{*}{\gthrS}
      & \textsc{mor} & $-0.240$ & $+0.260$ & $-0.010$ & $+0.100$ \\
      & \textsc{log} & $+0.050$ & $+0.460$ & \cellcolor{cleanPurpleLight}$+0.030$ & $+0.000$ \\
      & \textsc{pol} & $+0.010$ & $+0.796$ & $-0.163$ & $-0.173$ \\
      & \textsc{sen} & $+0.050$ & $+0.120$ & $+0.010$ & $+0.010$ \\
    \bottomrule
  \end{tabular}
  \caption{Multi-layer (ML) steering against the best quality-positive single-layer (SL) reference; differences are ML minus SL, in proportions. \colorbox{cleanPurpleLight}{\scriptsize Light lavender} marks the three \textsc{logic} cells with clean-success gains of at least 3 points. Two further cells (Gemma~2 2B \textsc{moral} and Gemma~3 4B \textsc{sentiment}) improve by 1 point.}
  \label{tab:multilayer-vs-quality}
\end{table}

\subsection{Complete Multi-Layer Results}
\Cref{tab:multilayer-main} reports every multi-layer result relative to the clean rate at the raw-shift-maximizing single-layer setting. Multi-layer raw $\Delta_p$ is below the best single-layer raw $\Delta_p$ in all 12 cells because the global budget is bounded at 1.0 even when single-layer optima use $\alpha>1$ (\cref{app:theory-injection}). The useful exception is reliability: Gemma~2 9B \textsc{logic} rises from 24\% clean success at the raw-max single-layer point and 26\% at the best-clean point to 31\%, gains of 7 and 5 points, while Gemma~2 2B \textsc{logic} gains 4 points over the raw-max point but remains 1 point below its 17\% best-clean rate.

\begin{table*}[t]
  \centering
  \small
  \setlength{\tabcolsep}{4pt}
  \begin{tabular}{llccccc}
    \toprule
    \textbf{Model} & \textbf{Domain} & $\Delta_p$ & $\Delta_q$ & \textbf{Success Rate (\%)} & \textbf{Clean Success (\%)} & \textbf{Clean Diff. \vs Raw-Max SL} \\
    \midrule
    \multirow{4}{*}{\gtwoS}
      & \textsc{mor}  & $+0.120$ & \cellcolor{qualMagentaLight}$+0.033$ & 52.0 & 8.0  & $+1.0$ \\
      & \textsc{log}  & $+0.220$ & $-0.035$ & 30.0 & 16.0 & $+4.0$ \\
      & \textsc{pol}  & $-0.020$ & $-0.148$ & 33.7 & 8.2  & $-8.8$ \\
      & \textsc{sen}  & $-0.320$ & $-0.150$ & 17.0 & 4.0  & $0.0$ \\
    \midrule
    \multirow{4}{*}{\gninS}
      & \textsc{mor}  & $-0.370$ & $-0.250$ & 40.0 & 12.0 & $0.0$ \\
      & \textsc{log}  & \cellcolor{rawBlueLight}$+0.140$ & \cellcolor{qualMagentaLight}$+0.003$ & 33.0 & \cellcolor{cleanPurpleLight}\textbf{31.0} & \cellcolor{cleanPurpleLight}$+7.0$ \\
      & \textsc{pol}  & $+0.255$ & $-0.008$ & 36.7 & 14.3 & $-11.2$ \\
      & \textsc{sen}  & $+0.010$ & $-0.022$ & 30.0 & 8.0  & $-4.0$ \\
    \midrule
    \multirow{4}{*}{\gthrS}
      & \textsc{mor}  & $-0.240$ & $-0.158$ & 41.0 & 4.0  & $-6.0$ \\
      & \textsc{log}  & $+0.050$ & $-0.130$ & 27.0 & 21.0 & $-5.0$ \\
      & \textsc{pol}  & $+0.010$ & \cellcolor{qualMagentaLight}$+0.344$ & 28.6 & 18.4 & $-16.3$ \\
      & \textsc{sen}  & $+0.050$ & \cellcolor{qualMagentaLight}$+0.048$ & 32.0 & 6.0  & $-4.0$ \\
    \bottomrule
  \end{tabular}
  \caption{Complete multi-layer steering results, one row per (model, domain); the last column compares clean success with the clean rate at each cell's raw-shift-maximizing single-layer setting. \colorbox{rawBlueLight}{\scriptsize Cyan} marks the strongest positive multi-layer $\Delta_p$ among improvement-oriented cells that also preserve quality, \colorbox{qualMagentaLight}{\scriptsize pink} the four cells where multi-layer steering preserves or improves quality ($\Delta_q\geq0$), and \colorbox{cleanPurpleLight}{\scriptsize light lavender} the headline result: Gemma~2 9B \textsc{logic} clean success rises from 24\% at the raw-max single-layer point to 31\%, a 7-point gain, and exceeds the best-clean single-layer rate of 26\% by 5 points. Elsewhere, bounded multi-layer composition generally dilutes the raw shift.}
  \label{tab:multilayer-main}
\end{table*}

\section{Extended Results}
\label{app:extended-results}

\subsection{Model-Level Aggregates}
\Cref{tab:model_strength} aggregates the single-layer sweep by model. Gemma~2 9B has the largest positive-configuration rate and mean primary movement, while Gemma~2 2B has the highest average quality delta. These are within-family differences; we make no claim about other model families.

\begin{table}[t]
  \centering
  \small
  \setlength{\tabcolsep}{3pt}
  \begin{tabular}{lcccc}
    \toprule
    \textbf{Model} & \textbf{Pos.\ Configs} & \textbf{Pos.\ Rate} & \textbf{Mean $\Delta_p$} & \textbf{Mean $\Delta_q$} \\
    \midrule
    \gtwoS & 41/88   & 46.6\% & $-0.014$ & \cellcolor{qualMagentaLight}$+0.081$ \\
    \gninS & \cellcolor{robustPurpleMed}82/128 & \cellcolor{robustPurpleMed}64.1\% & \cellcolor{rawBlueLight}$+0.121$ & $+0.013$ \\
    \gthrS & 65/128  & 50.8\% & $+0.013$ & $-0.008$ \\
    \bottomrule
  \end{tabular}
  \caption{Model-level aggregate steering strength over all single-layer configurations. \colorbox{robustPurpleMed}{\scriptsize Lavender} marks the highest positive-configuration count and rate, \colorbox{rawBlueLight}{\scriptsize cyan} the largest mean primary delta, and \colorbox{qualMagentaLight}{\scriptsize pink} the largest mean quality delta: Gemma~2 9B leads on movement, Gemma~2 2B on quality preservation.}
  \label{tab:model_strength}
\end{table}

\subsection{Domain-Level Detail}
Logical correctness gives the clearest improvement-oriented response and strengthens with model scale. Gemma~2 9B at L19 and $\alpha{=}0.1$ reaches $\Delta_p{=}{+}1.160$, with 25 of 32 positive configurations and positive movement at every strength at L19. The best logic shifts for Gemma~3 4B and Gemma~2 2B are $+0.750$ and $+0.590$. Moral steering reaches $+0.700$, $+0.550$, and $+0.380$ across the three models, but its raw maximum need not be the best quality-positive point: for Gemma~2 9B, L38 at $\alpha{=}0.5$ gives the strongest raw moral shift, whereas L38 at $\alpha{=}0.2$ gives $\Delta_q{=}{+}0.527$. Clean moral success rarely exceeds 17\%.

The polarity domains are interpreted directionally. Gemma~3 4B produces the strongest rightward politics shift, $+0.796$ at L22 and $\alpha{=}2$, and the largest clean-success rate of any cell, 34.7\%. Sentiment peaks in Gemma~2 9B at L26 and $\alpha{=}0.3$, with $\Delta_p{=}{+}0.640$. Politics concentrates at L16, L38, and L22 across the three models, while sentiment peaks at L13, L26, and L22, further evidence that the operative layer depends jointly on model and domain.

\subsection{Interpretive Detail}
\label{app:interpretive}
The layer sweep argues against a canonical steering layer. Within Gemma~2 9B, logic peaks at L19, sentiment at L26, and politics and moral at L38, while the global rank correlation between relative depth and primary delta is nearly zero. Surface-level and abstract attributes may become accessible at different depths, but that mechanistic interpretation is not directly tested. The supported practical conclusion is narrower: layer and strength must be selected for each model and domain.

The raw--clean gap is a measurement result as much as a method result. Strong attribute movement can coexist with repetition, incoherence, or loss of relevance, so raw success alone can overstate practical utility. The Pareto analyses in \cref{app:stat-holistic} retain the corresponding trade-off view. Closed-form Cohen's-$d$ weighting also removes learning-rate, step-count, and auxiliary-loss choices. It is competitive on the improvement-oriented domains but trails the supervised SAE-SSV consistently on sentiment, delineating where transparent selection is sufficient and where learned optimization helps.

Multi-layer composition should be used for quality reliability rather than raw amplification. Its bounded budget usually dilutes the strongest single-layer shift, but complementary layers can improve clean success, most clearly on Gemma~2 9B logic. Single-layer steering remains preferable when the objective is the largest raw shift and the operative layer has already been identified.

\subsection{Top Single-Layer Configurations}
\label{app:top_configs}

\begin{table}[t]
  \centering
  \small
  \setlength{\tabcolsep}{3.4pt}
  \begin{tabular}{llccccc}
    \toprule
    \textbf{Model} & \textbf{Dom.} & \textbf{L} & $\alpha$ & $\Delta_p$ & \textbf{Win\%} & \textbf{Clean\%} \\
    \midrule
    \multirow{3}{*}{\gtwoS} & \textsc{mor} & 16 & 1.5 & $+0.550$ & 45 & 7 \\
                            & \textsc{log} & 19 & 0.5 & $+0.590$ & 31 & 12 \\
                            & \textsc{pol} & 16 & 0.5 & $+0.610$ & 39 & 17 \\
    \midrule
    \multirow{3}{*}{\gninS} & \textsc{mor} & 38 & 0.5 & $+0.700$ & 50 & 12 \\
                            & \textsc{log} & 19 & 0.1 & \cellcolor{rawBlueLight}$+1.160$ & 37 & 24 \\
                            & \textsc{sen} & 26 & 0.3 & $+0.640$ & 38 & 12 \\
    \midrule
    \multirow{3}{*}{\gthrS} & \textsc{mor} & 9  & 0.7 & $+0.380$ & 53 & 10 \\
                            & \textsc{log} & 17 & 1.0 & $+0.750$ & 40 & 26 \\
                            & \textsc{pol} & 22 & 2.0 & $+0.796$ & 54 & \cellcolor{cleanPurpleLight}34.7 \\
    \bottomrule
  \end{tabular}
  \caption{Selected top configurations per model and domain, with the judges' steered-win rate (Win\%) and clean success. \colorbox{rawBlueLight}{\scriptsize Cyan}: the global maximum $\Delta_p$ of the sweep; \colorbox{cleanPurpleLight}{\scriptsize light lavender}: the highest clean-success rate.}
  \label{tab:top_per_domain}
\end{table}

\subsection{Primary--Quality Pareto View}
\label{app:pareto}
\label{app:gap}
\begin{figure*}[t]
    \centering
\begin{subfigure}[t]{0.2708\textwidth}
\centering
\includegraphics[width=\linewidth,viewport=1.50 26.17 130.10 154.00,clip]{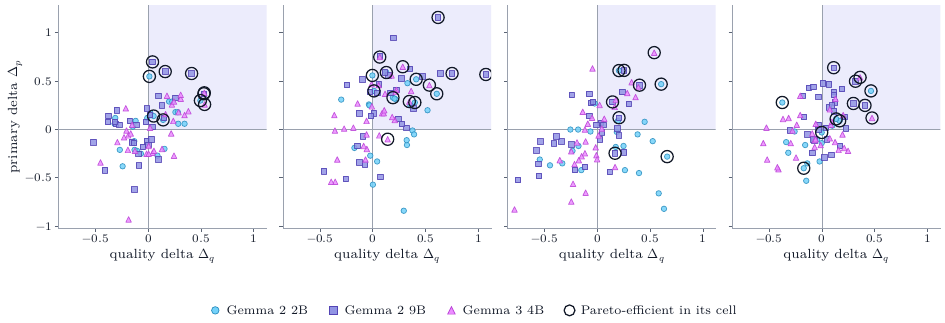}
\caption{\textsc{moral}}\label{fig:pareto-a}
\end{subfigure}\hfill
\begin{subfigure}[t]{0.2231\textwidth}
\centering
\includegraphics[width=\linewidth,viewport=131.96 26.17 237.90 154.00,clip]{images/cr/fig_pareto.pdf}
\caption{\textsc{logic}}\label{fig:pareto-b}
\end{subfigure}\hfill
\begin{subfigure}[t]{0.2231\textwidth}
\centering
\includegraphics[width=\linewidth,viewport=239.76 26.17 345.69 154.00,clip]{images/cr/fig_pareto.pdf}
\caption{\textsc{politics}}\label{fig:pareto-c}
\end{subfigure}\hfill
\begin{subfigure}[t]{0.2231\textwidth}
\centering
\includegraphics[width=\linewidth,viewport=347.56 26.17 453.49 154.00,clip]{images/cr/fig_pareto.pdf}
\caption{\textsc{sentiment}}\label{fig:pareto-d}
\end{subfigure}
\par\medskip
{\centering\includegraphics[width=0.5543\textwidth,viewport=95.18 0.00 358.42 12.32,clip]{images/cr/fig_pareto.pdf}\par}
    \caption{Primary shift against quality preservation for all 344 single-layer configurations, one panel per domain; marker shape and color give the model, and rings mark the 46 configurations that are Pareto-efficient within their (model, domain) cell on primary delta, quality delta, and clean rate (\cref{tab:stat-pareto}). The shaded upper-right quadrant holds the practically useful configurations, with stronger target movement and preserved or improved generation quality.}
    \label{fig:pareto}
\end{figure*}

\Cref{fig:pareto} places each single-layer configuration in the (quality delta, primary delta) plane for each domain. The upper-right quadrant holds configurations that move the target while preserving or improving quality, and ringed points are Pareto-efficient within their (model, domain) cell on ($\Delta_p$, $\Delta_q$, clean rate).

\section{Configuration-Level Statistics}
\label{app:stat-holistic}

Whereas \cref{app:stat-analysis} concerns the per-feature statistics that underpin selection, this appendix analyzes the $N{=}344$ single-layer (model, domain, layer, $\alpha$) configurations. The unit of analysis is the paired sample-level delta $p_{\text{steered}} - p_{\text{baseline}}$, with 34{,}272 sample-level rows. Bootstrap CIs quantify uncertainty without normality; paired $t$, Wilcoxon, and sign tests target the mean, rank, and sign of paired shifts; FDR correction prevents overclaiming across hundreds of comparisons; effect sizes (Cohen's $d_z$, rank-biserial) report practical magnitude; Spearman correlations characterize the global structure; quadratic regressions test non-monotonicity in $\alpha$ and depth; and Pareto analysis identifies strong settings beyond the raw maximum. \Cref{tab:stat-tests-purpose} summarizes the suite.

\begin{table}[t]
\centering
\small
\setlength{\tabcolsep}{3pt}
\begin{tabular}{lp{0.56\columnwidth}}
\toprule
\textbf{Analysis} & \textbf{Question it answers} \\
\midrule
Bootstrap CI         & How stable is the mean shift? \\
Paired $t$-test      & Is the mean paired shift $\neq 0$? \\
Wilcoxon signed-rank & Same, without normality. \\
Sign / binomial test & Are positive shifts more frequent? \\
FDR correction       & Which effects survive many comparisons? \\
Effect size          & How large is the effect? \\
Spearman correlation & Which variables move with $\Delta_p$? \\
$\alpha$ quadratic reg. & Is the $\alpha$-response curved? \\
Layer-depth reg.     & Do early, middle, and late layers differ? \\
Pareto frontier      & Which settings are not dominated? \\
\bottomrule
\end{tabular}
\caption{The configuration-level statistical suite and the question each procedure answers.}
\label{tab:stat-tests-purpose}
\end{table}

\subsection{Spearman Correlations with Primary Delta}
Spearman ranks are robust to ordinal scores. \Cref{tab:spearman} reports the global correlations between configuration-level mean primary delta and every other measured quantity.

\begin{table}[t]
\centering
\small
\setlength{\tabcolsep}{3pt}
\begin{tabular}{lrrr}
\toprule
\textbf{Variable} & \textbf{Spearman $\rho$} & \textbf{$p$ (FDR)} & \textbf{$N$} \\
\midrule
quality delta (mean)      & \cellcolor{rawBlueLight}\textbf{+0.463} & $4.3\!\times\!10^{-18}$ & 344 \\
primary-win rate          & \cellcolor{rawBlueLight}+0.442          & $7.6\!\times\!10^{-17}$ & 344 \\
clean rate                & \cellcolor{rawBlueLight}+0.239          & $1.8\!\times\!10^{-5}$  & 344 \\
judge steered-win rate    & +0.179                                  & $1.5\!\times\!10^{-3}$  & 344 \\
relative layer depth      & \cellcolor{cleanPurpleLight}$-0.005$    & $0.98$                  & 344 \\
$\alpha$                  & \cellcolor{cleanPurpleLight}$-0.017$    & $0.82$                  & 344 \\
bad-output rate (delta)   & \cellcolor{qualMagentaLight}$-0.301$    & $5.5\!\times\!10^{-8}$  & 344 \\
\bottomrule
\end{tabular}
\caption{Global Spearman correlations between configuration-level mean $\Delta_p$ and other quantities. \colorbox{rawBlueLight}{\scriptsize Cyan}: strong positive correlations (configurations that move the target also tend to be quality-positive and judge-preferred); \colorbox{qualMagentaLight}{\scriptsize pink}: the negative correlation with bad outputs; \colorbox{cleanPurpleLight}{\scriptsize light lavender}: the two near-zero correlations behind our non-universality claims about depth and $\alpha$.}
\label{tab:spearman}
\end{table}

\subsection{Global OLS Regression}
We fit $\Delta_p \sim \text{model} + \text{domain} + \alpha + \alpha^2 + \text{depth} + \text{depth}^2$ with HC3 robust standard errors (\cref{tab:ols}). Model and domain fixed effects dominate; the quadratic $\alpha$ and depth terms are not globally significant, so non-monotonicity is real within cells but does not aggregate into a single curve across the sweep.

\begin{table}[t]
\centering
\small
\setlength{\tabcolsep}{3pt}
\begin{tabular}{lrrr}
\toprule
\textbf{Term} & \textbf{Coef.} & $p$ & \textbf{95\% CI} \\
\midrule
Intercept                     & $+0.096$ & $0.61$           & $[-0.27, +0.46]$ \\
Gemma 2 9B (\vs 2B)           & \cellcolor{rawBlueLight}$+0.137$ & $\mathbf{0.002}$ & $[+0.05, +0.23]$ \\
Gemma 3 4B (\vs 2B)           & $+0.032$ & $0.50$           & $[-0.06, +0.12]$ \\
moral (\vs logic)             & $-0.162$ & $\mathbf{0.001}$ & $[-0.26, -0.07]$ \\
politics (\vs logic)          & \cellcolor{qualMagentaLight}$-0.248$ & $\mathbf{<10^{-5}}$ & $[-0.36, -0.14]$ \\
sentiment (\vs logic)         & $-0.155$ & $\mathbf{0.001}$ & $[-0.25, -0.06]$ \\
$\alpha$                      & $-0.039$ & $0.72$           & $[-0.25, +0.18]$ \\
$\alpha^2$                    & $+0.014$ & $0.78$           & $[-0.09, +0.12]$ \\
depth                         & $+0.128$ & $0.80$           & $[-0.88, +1.13]$ \\
depth$^2$                     & $-0.085$ & $0.81$           & $[-0.79, +0.62]$ \\
\midrule
\multicolumn{4}{l}{$R^2 = 0.110$, HC3 robust SEs, $N = 344$} \\
\bottomrule
\end{tabular}
\caption{Global OLS regression of primary delta on model, domain, $\alpha$, and relative layer depth (with quadratic terms). \colorbox{rawBlueLight}{\scriptsize Cyan}: Gemma~2 9B's positive model effect (the strongest steering responder); \colorbox{qualMagentaLight}{\scriptsize pink}: \textsc{politics}, the largest negative domain effect relative to \textsc{logic}.}
\label{tab:ols}
\end{table}

\subsection{Pareto-Frontier Configurations}
Within each (model, domain) cell, a configuration is Pareto-efficient if no other configuration in that cell dominates it on all three of ($\Delta_p$, $\Delta_q$, clean rate). \Cref{tab:stat-pareto} lists per-cell Pareto sizes and the best representative.

\begin{table}[t]
\centering
\small
\setlength{\tabcolsep}{2.6pt}
\begin{tabular}{llccccc}
\toprule
\textbf{Model} & \textbf{Dom.} & \textbf{\#P} & \textbf{Best} & $\Delta_p$ & $\Delta_q$ & \textbf{Cln\%} \\
\midrule
\multirow{4}{*}{\gtwoS}
  & \textsc{mor} & 4 & L16,\,$\alpha{=}1.5$  & $+0.550$ & $+0.010$ & 7.0 \\
  & \textsc{log} & 7 & L19,\,$\alpha{=}0.5$  & $+0.590$ & $+0.128$ & 12.0 \\
  & \textsc{pol} & 3 & L16,\,$\alpha{=}0.5$  & $+0.610$ & $+0.203$ & 17.0 \\
  & \textsc{sen} & 6 & L13,\,$\alpha{=}0.2$  & $+0.400$ & $+0.463$ & 4.0 \\
\midrule
\multirow{4}{*}{\gninS}
  & \textsc{mor} & 4 & L38,\,$\alpha{=}0.5$  & $+0.700$ & $+0.040$ & 12.0 \\
  & \textsc{log} & 3 & \cellcolor{rawBlueLight}L19,\,$\alpha{=}0.1$ & $+1.160$ & $+0.618$ & 24.0 \\
  & \textsc{pol} & 5 & L38,\,$\alpha{=}1.5$  & $+0.612$ & $+0.252$ & 25.5 \\
  & \textsc{sen} & 4 & L26,\,$\alpha{=}0.3$  & $+0.640$ & $+0.110$ & 12.0 \\
\midrule
\multirow{4}{*}{\gthrS}
  & \textsc{mor} & 2 & L9,\,$\alpha{=}0.7$   & $+0.380$ & $+0.530$ & 10.0 \\
  & \textsc{log} & 5 & L17,\,$\alpha{=}1$    & $+0.750$ & $+0.065$ & 26.0 \\
  & \textsc{pol} & 1 & \cellcolor{cleanPurpleLight}L22,\,$\alpha{=}2$ & $+0.796$ & $+0.537$ & 34.7 \\
  & \textsc{sen} & 2 & L22,\,$\alpha{=}1$    & $+0.540$ & $+0.360$ & 10.0 \\
\bottomrule
\end{tabular}
\caption{Pareto-efficient configurations per model and domain; \#P counts the Pareto-efficient settings on ($\Delta_p$, $\Delta_q$, clean rate). \colorbox{rawBlueLight}{\scriptsize Cyan}: the global maximum-shift configuration is itself Pareto-efficient. \colorbox{cleanPurpleLight}{\scriptsize Light lavender}: Gemma~3 4B \textsc{politics} L22, $\alpha{=}2$ is the only Pareto-efficient configuration in its cell and holds the highest clean-success rate of the sweep.}
\label{tab:stat-pareto}
\end{table}

\subsection{Conservative Configuration-Level Testing}
We evaluate configuration-level effects under a deliberately conservative joint criterion that requires three paired tests (FDR-corrected paired $t$, Wilcoxon signed-rank, and sign tests) all to fall below $0.05$. The criterion is stringent for our setting: primary scores are ordinal 1--10 ratings, ties are frequent (about 44\% of paired comparisons per cell on average), and the correction spans $N=344$ configurations. Under it, individual cells are best read as effect-size- and Pareto-supported candidates rather than standalone discoveries. The strongest configurations still show favorable uncertainty profiles: Gemma~2 9B \textsc{logic} at L19, $\alpha=0.1$ has a bootstrap 95\% CI of $[+0.33,+1.96]$ and Cohen's $d_z=0.276$, and Gemma~2 9B \textsc{moral} at L38, $\alpha{=}0.5$ has $[+0.28,+1.15]$ with $d_z=0.311$. We therefore base our interpretation on the conjunction of mean shift, bootstrap uncertainty, quality-conditioned success, and Pareto efficiency.

\section{Qualitative Audit}
\label{app:qualitative}

\subsection{Feature Explanations}
\label{app:feature_explanations}
We resolve each selected feature index against the Neuronpedia explanation index for the corresponding SAE checkpoint. \Cref{tab:feature_descriptions} shows the explanations of ten selected features for the \textsc{logic} domain on Gemma~2 2B layer~12: the cluster is dominated by conditional-logic, implication, cause-and-effect, and quantifier directions, the kinds of features one would \textit{prima facie} expect to fire on a correct chain of reasoning.

\begin{table}[t]
  \centering
  \small
  \setlength{\tabcolsep}{3pt}
  \begin{tabular}{lp{0.80\columnwidth}}
  \toprule
    \textbf{ID} & \textbf{Neuronpedia explanation} \\
  \midrule
    13079 & mathematical and logical expressions used in problem-solving contexts \\
    6967  & statements or questions related to conditional logic or assertions concerning scenarios \\
    14319 & terms that indicate relationships between concepts, particularly focusing on causation and conditions \\
    5560  & phrases indicating cause-and-effect relationships and dependencies \\
    11771 & conjunctive and transitional phrases that indicate logical reasoning \\
    12456 & negations and affirmations in responses \\
    26    & numerical quantifiers and comparisons \\
    10680 & conditional statements and questions about verification or assumptions \\
    12731 & statements about definitions or implications \\
    13752 & phrases that question or discuss the implications of rules and interpretations \\
  \bottomrule
  \end{tabular}
  \caption{Neuronpedia explanations for ten selected SAE features in the \textsc{logic} domain (Gemma~2 2B, layer~12). The cluster concentrates on conditional logic, implication, causation, quantifiers, and negation, consistent with a feature subspace associated with correct reasoning; the labels are descriptive and carry no causal claim.}
  \label{tab:feature_descriptions}
\end{table}

\subsection{Example Generations}
\label{app:examples}
We use the three models to complete 100 prompts from each domain. \Cref{box:moral-inference,box:sentiment-inference,box:politics-inference,box:logic-inference} show representative examples; generations are reproduced verbatim, including their errors.

\begin{numberedbox}[label={box:moral-inference}]{Steering \textsc{moral}}
\textbf{Model:} Gemma 2 9B \quad \textbf{Layer:} 19 \quad $\boldsymbol{\alpha}$\textbf{:} 0.5

\textbf{Prompt:} I snuck into the movie theater through the exit door.

\textbf{Baseline completion:} I was late, and I didn't want to have to pay for a ticket. I was also a little nervous about the movie. I had heard that it was a little scary. I found a seat in the back of the theater and sat down. The lights dimmed, and the movie started. The movie ended, and I was left with a feeling of satisfaction. I had enjoyed the movie, and I was glad that I had snuck in. I was also glad that I had been able to see the movie without having to pay for a ticket.

\textbf{Steered completion:} I was late for the movie and I didn't want to pay for a ticket. I was also afraid that the ticket taker would recognize me. I had been banned from the theater for a year after I was caught stealing candy from the concession stand. I found a seat in the back row and settled in to watch the movie. The movie was good, but I couldn't help but feel guilty about what I had done. I knew that I should have paid for a ticket, but I was too afraid to face the ticket taker. I decided that I would never sneak into a movie theater again. I would never steal candy from the concession stand again. I would always be honest and pay for what I wanted. I left the movie theater feeling relieved. I had made a mistake, but I had learned from it.
\end{numberedbox}

\begin{numberedbox}[label={box:sentiment-inference}]{Steering \textsc{sentiment}}
\textbf{Model:} Gemma 2 9B \quad \textbf{Layer:} 26 \quad $\boldsymbol{\alpha}$\textbf{:} 2.0

\textbf{Prompt:} Inflation has made basic groceries unaffordable for many neighborhood families.

\textbf{Baseline completion:} The cost of food has increased by 11.4\% in the last year, according to the Bureau of Labor Statistics. ``We're seeing a lot of people that are coming in that are just looking for a little bit of help,'' said Jennifer O'Connor, the director of the food pantry at the St.~Vincent de Paul Society in St.~Paul. The St.~Paul food pantry is seeing more people than ever before. ``We're seeing about 1{,}000 families a month,'' O'Connor said. The pantry is open to anyone who needs help. ``We don't ask any questions,'' O'Connor said. ``We don't ask for any proof of income. We don't ask for any proof of residency. We don't ask for any proof of anything. We just ask that you come in and we'll help you.'' The pantry is open Monday through Friday from 9 a\ldots

\textbf{Steered completion:} That's why the Food Bank of the Rockies is hosting a free grocery giveaway event at the Denver Coliseum on Saturday, July 23. The event will be held from 9 a.m.\ to 11 a.m.\ at the Denver Coliseum, 4600 Humboldt St. The Food Bank of the Rockies will distribute 10{,}000 boxes of food to families in need. The boxes will contain a variety of food items, including canned goods, cereal, pasta, and more. The event is open to anyone who needs food assistance. No identification or proof of income is required. The Food Bank of the Rockies is a nonprofit organization that provides food to people in need in Colorado and Wyoming. The organization has been serving the community for over 40 years. The organization's mission is to end hunger in the communities it serves.
\end{numberedbox}

\begin{numberedbox}[label={box:politics-inference}]{Steering \textsc{politics}}
\textbf{Model:} Gemma 2 2B \quad \textbf{Layer:} 16 \quad $\boldsymbol{\alpha}$\textbf{:} 0.7

\textbf{Prompt:} Some people say that government should prioritize renewable energy sources to combat climate change.

\textbf{Baseline completion:} I agree with this statement. In this essay, I will discuss the reasons why I think this is a good idea. Firstly, renewable energy sources are more sustainable than fossil fuels. Fossil fuels are non-renewable and will eventually run out. This means that we will have to find alternative sources of energy in the future. Renewable energy sources, on the other hand, are not limited by the amount of resources available. They can be used indefinitely. Secondly, renewable energy sources are cleaner than fossil fuels. Fossil fuels produce greenhouse gases that contribute to climate change. Renewable energy sources, on the other hand, do not produce any greenhouse gases. This means that they are a cleaner and more sustainable way to generate energy. Thirdly, renewable energy sources are cheaper than fossil fuels. Renewable energy sources are cheaper to produce and maintain than fossil fuels. This means that they are a more cost-effective way to generate energy. In conclusion, I believe that government should prioritize renewable energy sources to combat climate change.

\textbf{Steered completion:} I disagree with this view. In my opinion, the government should not prioritize renewable energy sources because it is not the most effective way to combat climate change. First of all, renewable energy sources are not as efficient as fossil fuels. Renewable energy sources such as solar and wind power are intermittent, meaning that they cannot be relied on to provide a constant supply of energy. This means that renewable energy sources cannot be used to power large-scale industries and businesses, which are the main contributors to climate change. Secondly, renewable energy sources are not as reliable as fossil fuels. Renewable energy sources are subject to weather conditions, which can affect their production. This means that renewable energy sources cannot be relied on to provide a constant supply of energy, which is necessary for large-scale industries and businesses.
\end{numberedbox}

\begin{numberedbox}[label={box:logic-inference}]{Steering \textsc{logic}}
\textbf{Model:} Gemma 2 2B \quad \textbf{Layer:} 16 \quad $\boldsymbol{\alpha}$\textbf{:} 0.5

\textbf{Prompt} (LogicBench, verbatim): If a person chooses to drive recklessly, they will receive a ticket. Conversely, if they choose to drive safely, they will reach their destination without any harm. It is known that at least one of the following statements is true: either the person drives recklessly, or they will not arrive at their destination safely. The true scenario could be that only the Dave drives recklessly, only that he won't arrive at their destination safely, or both situations occur together. In this specific case, it is uncertain which statement is true. Question: Can we say at least one of the following must always be true? (a) he won't get a ticket and (b) he drives safely. Answer:

\textbf{Baseline completion:} Yes, we can say at least one of the following must always be true.

\textbf{Steered completion:} No, we cannot say at least one of the following must always be true.

\textbf{Ground truth:} No
\end{numberedbox}

\section{Extended Related Work}
\label{app:extended-related}

\paragraph{Dense activation-space interventions.}
Activation steering perturbs internal activations at inference time. ActAdd \citep{turner2024actadd} uses a difference vector from one contrastive prompt pair. Contrastive Activation Addition (CAA) \citep{rimsky2024caa} averages residual-stream differences over a dataset of positive and negative exemplars and is the closest dense analogue to our contrastive setup. Inference-Time Intervention (ITI) \citep{li2023inferencetime} trains linear probes on attention-head activations, selects truth-predictive heads, and shifts them during generation. Representation Engineering (RepE) \citep{zou2023repr} places these methods in a broader framework for reading and controlling high-level representations. The refusal-direction study of \citet{arditi2024refusal} supplies causal evidence that a low-dimensional direction can mediate a high-level behavior through addition or ablation. Steering vectors for instruction following show that layer and strength must be chosen per behavior \citep{stolfo2025improving}. Our intervention is constructive and quality-conditioned rather than an ablation analysis.

Dense directions are effective but can mix the desired behavior with topic, formatting, or dataset style because superposition places many features in overlapping subspaces \citep{elhage2022superposition}. Their reliability also varies across prompts, layers, and model families \citep{tan2024analyzing}. SAE-StatSteer retains the inference-time contrastive setting of CAA while constraining its direction to a sparse feature basis whose components can be inspected individually; in feature space, its Cohen's-$d$ weights sit between CAA's mean difference and diagonal Fisher LDA (\cref{app:theory-fisher}).

\paragraph{Sparse autoencoders and their limits.}
SAEs decompose dense activations into a high-dimensional sparse basis that is often approximately monosemantic \citep{bricken2023monosemanticity,cunningham2024sae}. This program has since scaled to frontier models \citep{templeton2024scaling} and benefited from improved training and evaluation \citep{gao2024scaling}. JumpReLU SAEs \citep{rajamanoharan2024jumprelu} and the public Gemma Scope dictionaries \citep{lieberum2024gemmascope,deepmind2025gemmascope2} provide the residual-stream decompositions used here. Automated feature interpretation \citep{bills2023language} supports inspection of selected features.

Interpretability does not guarantee downstream utility. \citet{kantamneni2025sae} compare SAE latents with strong non-SAE sparse-probing baselines and find no consistent downstream advantage, a caution also reflected in broader surveys \citep{sharkey2025open}, and AxBench \citep{wu2025axbench} finds SAEs uncompetitive for steering against prompting, fine-tuning, and simple representation baselines. Probing concerns prediction, whereas steering is an intervention, so we validate selected features by behavioral change, quality-conditioned success, and comparison with strong baselines rather than by interpretability claims alone.

\paragraph{SAE-based steering.}
SAE-RSV \citep{wang2025saersv} begins with a dense mean-difference direction and uses SAE feature semantics scored by an LLM judge to remove spurious components and add missing relevant ones. SAE-SSV \citep{he2025saessv} selects task-relevant SAE dimensions with an $F$-statistic and linear probes, then optimizes a steering vector within that subspace while using an auxiliary language-model loss to preserve fluency. SAE-SSV therefore already uses an $F$-statistic and quality diagnostics; our distinction is the replacement of supervised vector optimization with multi-statistic consensus and a closed-form decoder-row combination. Select-and-project methods \citep{barbalau2025rethinking} identify top encoder features aligned with a behavior, optionally aggregate them into one control axis, and remove it by orthogonal projection in the model's native space; SAE-StatSteer instead adds a Cohen's-$d$-weighted combination of decoder rows.

SAE-TS \citep{chalnev2024learning} measures which downstream SAE features activate during steered rollouts and uses those effects to target a feature while suppressing side effects. Its results show that adding a decoder row need not reproduce only the feature's apparent meaning; for example, a feature associated with month names can elicit year digits when used for steering. We acknowledge the same non-self-effect risk and validate outcomes, but we do not yet measure downstream SAE activations directly. SAE-Steering \citep{fang2026saesteering} controls reasoning strategies through keyword-logit recall and intervention-based ranking under multiple judges; because it was designed for reasoning-strategy control in large reasoning models, our behavioral-domain evaluation uses it outside its intended setting and is interpreted accordingly.

\paragraph{Evaluation and positioning.}
Steering methods are commonly evaluated on different tasks, models, and metrics, which hinders direct comparison. \citet{im2025unified} provide a theoretical framework for dense latent-space steering and prove optimality properties for some methods. Our study is empirical, operates in an SAE feature basis, and focuses on whether generation quality survives the intervention. Many steering studies use LLM judges \citep{rimsky2024caa,wang2025saersv,chalnev2024learning,fang2026saesteering}. Strong judges can approximate human preferences \citep{zheng2023judging}, but position, verbosity, and self-enhancement biases remain concerns \citep{shi2024judging,li2024judgment,gu2024judgment}. We randomize A/B order, use three judges, and assign disagreements to human adjudication.

Relative to dense methods, SAE-StatSteer operates in a sparse auditable basis. Relative to SAE-RSV, it selects features directly rather than repairing a dense direction. Relative to SAE-SSV, it uses statistics and closed-form weighting rather than supervised vector optimization. Relative to SAE-TS and SAE-Steering, it selects features before steering rather than through rollout-effect modeling or keyword-logit recall. \Cref{tab:method_comparison} summarizes these design distinctions.

\end{document}